# An Integrated Vision-and-Language Pretraining (VLP) and Visual Question Answering (VQA) model to Automate Nondestructive Evaluation Image Analysis

Mehrdad S. Dizaji, Ph.D.
Post-Doctoral Research Fellow, Federal Highway Administration,
Turner-Fairbank Highway Research Center, McLean, VA 22101,
m.shafiei.dizaji.ctr@dot.gov; mehrdadshafiei@gmail.com

Hoda Azari, Ph.D.
Nondestructive Evaluation Program and Laboratory Manager,
Federal Highway Administration,
Turner-Fairbank Highway Research Center, McLean, VA 22101,
hodaazari@dot.gov

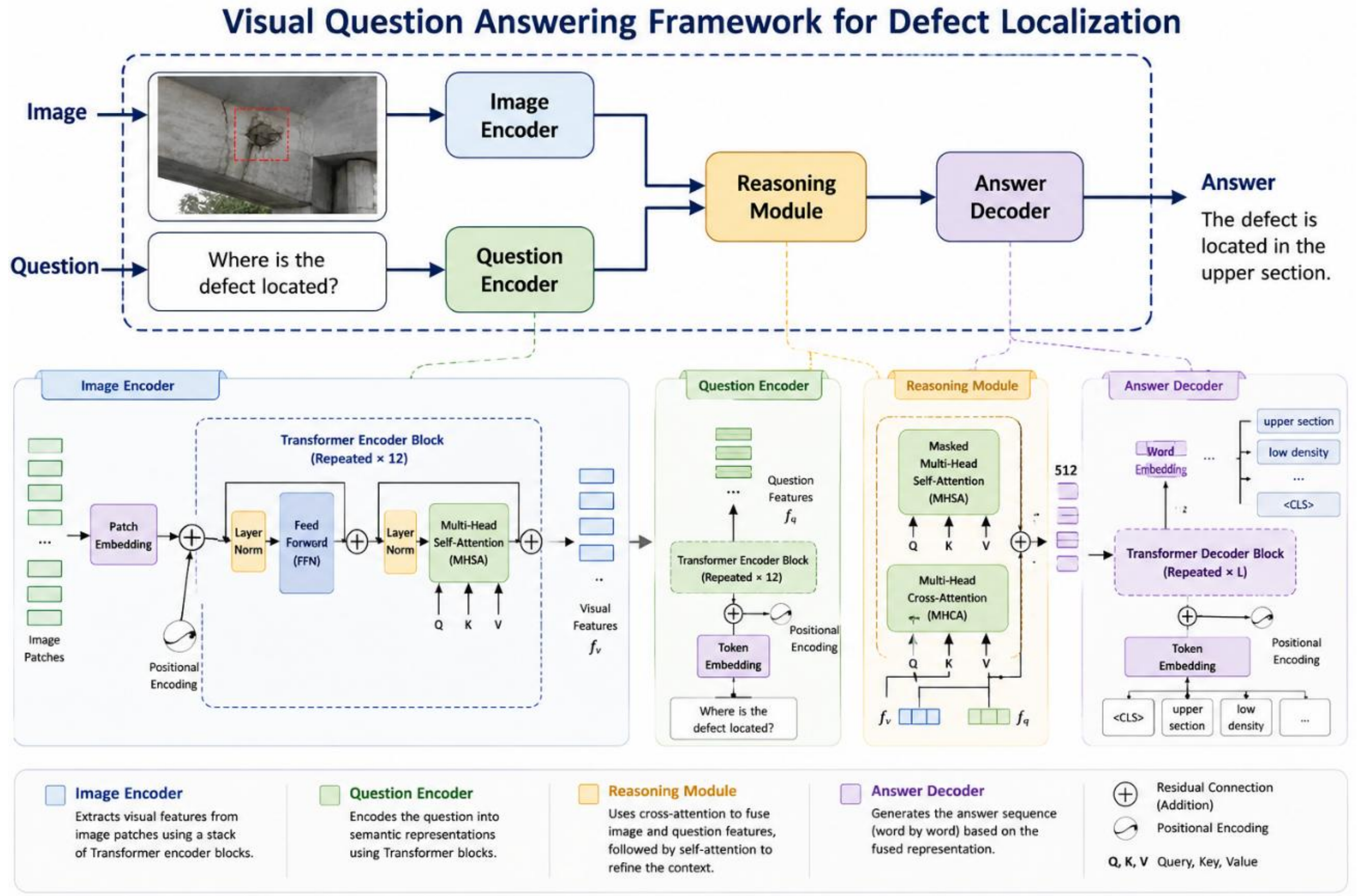

**Summary**
Nondestructive evaluation (NDE) methods—like ultrasonic testing, radiographic imaging, and thermography—are commonly used to check the health of materials and infrastructure without causing any damage. Making sense of those images usually takes a lot of time and expert knowledge, which makes the whole process slower and more expensive than it needs to be. That's where this study comes in. We're introducing an AI-based approach called ChatNDE-Figure-to-Caption, which aims to automate the interpretation of NDE images using deep learning and natural language processing (NLP). The idea is to take a load off inspectors by training a model that can look at an NDE image and generate a detailed, accurate caption that explains what's going on. We're using a Vision-and-Language Pretraining (VLP) strategy to help the model learn how to connect visual features with meaningful language. Basically, we built a large NDE image dataset, trained the model using annotated examples, and then evaluated how well it performed using BLEU scores to compare its output to expert-written descriptions. So, the system combines a ResNet-50 model to extract important features from the images and a GPT-2 language model to turn those features into natural-sounding text. Even though the accuracy of the model has been low the generated caption results have been solid so far—the captions were shorter but mentioned some important features of images what human experts would say, which shows the model is learning to pick up on key details. To make it easy to use, we're also building a user-friendly interface where people can upload NDE images and instantly get captioned feedback. There's even a feedback option built in, so users can help fine-tune the model over time. On top of that, we're looking into using a Visual Question Answering (VQA) model as part of the system. VQA models are designed to take an image and a question about that image (like "Is there a crack?" or "Where is the defect located?") and generate a useful answer. By adding this layer, the platform won't just describe what it sees—it can also respond to specific questions, making it even more interactive and helpful for inspectors in the field. This whole approach is a big step toward speeding up NDE workflows, reducing human error, and making the technology more accessible. Moving forward, we plan to grow the dataset, explore multimodal fusion techniques (combining different types of NDE data), and fine-tune the model to handle a wider range of inspection scenarios.

## 1. Introduction

Non-Destructive Evaluation (NDE) [1] plays a crucial role in assessing the structural integrity, safety, and longevity of materials used in various engineering applications, particularly in infrastructure, transportation, and construction industries. Think of bridges, buildings, tunnels, and pipelines—these things go through a lot over time, so it's crucial to catch any hidden issues before they turn into serious problems. NDE helps engineers and inspectors figure out what's going on inside materials without having to cut them open or damage them, which is obviously a huge plus [2]. Spotting things like cracks, voids, or other internal flaws early on can save a ton of money and, more importantly, can help prevent accidents and protect lives [3]. That said, the way NDE is traditionally done can be kind of a pain. It usually takes a lot of hands-on work, and you need highly trained people to interpret the results correctly. It's not just about taking a scan or an image—it's about knowing what you're looking at and being able to tell if something's off. That level of skill takes time to develop, and even then, human judgment isn't perfect. Mistakes can happen, and going through all the data manually can be slow and expensive [4]. That's where technology—and specifically artificial intelligence, starts to look really appealing.

Lately, there's been a lot of interest in using AI to help out with NDE, and for good reasons. Tools like deep learning and natural language processing (NLP) [5] are changing the game by making it possible to automate some of the trickiest parts of the process. For example, deep learning models can be trained to recognize patterns and features in images that point to defects or abnormalities, even the subtle ones that might be easy to miss. On top of that, NLP can be used to automatically generate clear, human-readable explanations of what the AI is seeing, which is super helpful for engineers and decision-makers [6]. That's exactly what the ***NDE_Chat*** Platform is aiming to do. It's an AI-powered system built to make analyzing NDE images a whole lot easier and more reliable. It uses state-of-the-art deep learning models—like convolutional neural networks (CNNs) to break down visual data and pull-out important features, and transformer-based language models to turn those insights into written descriptions. So instead of spending hours going over every single scan or trying to explain findings manually, the platform can handle a lot of that heavy lifting automatically. The end goal here is to reduce the workload on inspectors, cut down on human error, and give people more confidence in the results. By speeding up the analysis and making it more consistent, the platform can help teams catch issues earlier, understand material conditions better, and make smarter decisions when it comes to maintenance and repairs. It's all about combining the strengths of AI with the knowledge of experienced professionals to build safer, longer-lasting structures in a faster and more efficient way.

## 2. Objectives

The main goal of this research is to make the process of analyzing NDE (Nondestructive Evaluation) images faster, easier, and more reliable by using AI. Right now, interpreting these kinds of images—whether they come from Electrical Resistivity (ER), Ground Penetrating Radar (GPR), Impact Echo (IE), or Ultrasonic Surface Waves (USW)—takes a lot of time and often depends on the experience and judgment of individual inspectors. These techniques are super useful for understanding what's going on beneath the surface of materials and structures but manually going through the data can be slow and sometimes inconsistent. So, the first thing we're tackling is building an AI system that can automatically and accurately analyze all these different types of NDE images. By training deep learning models on large sets of labeled examples, the goal is to create a tool that can recognize patterns and detect issues with way more consistency than traditional methods. This should help cut down on subjective interpretations and standardize the whole process. The second part of the project is all about making the results more accessible. Instead of relying on technical reports full of jargon, we're working on generating clear, useful captions for each image. These captions are designed to summarize what's going on in simple, natural language—making it easier for engineers and inspectors to quickly understand the findings and take action when needed. The third objective is to boost the speed and accuracy of all this. By using advanced deep learning techniques, the system is being built to handle large amounts of data quickly, giving users real-time feedback instead of making them wait. To make the platform even more interactive, we're also incorporating Vision-Language Pretraining (VLP) [7] and Visual Question Answering (VQA) [8] models. VLP helps the AI understand the connection between images and language, while VQA allows users to ask specific questions about an image—like "Is there a void present?" or "What type of defect is shown?"—and get clear answers. Altogether, this project is shaping up to be a powerful AI-powered platform that helps streamline infrastructure inspections, supports smarter maintenance planning, and ultimately contributes to safer and more efficient decision-making in the field.

## 3. Phase 1: Vision-Language Pretraining (VLP) for Figure-to-Caption Generation

### *3.1 Methodology*

The first phase of this research focuses on the development of a Vision-Language Pretraining (VLP) model designed to generate high-quality, contextually accurate captions for NDE images. The whole picture of this phase is shown in Figure 1. This phase consists of multiple stages, including data collection, preprocessing, model training, and performance evaluation. The data collection process involves compiling a comprehensive dataset of NDE images sourced from various imaging techniques, ensuring a diverse and representative dataset that captures a wide range of structural conditions, material defects, and subsurface anomalies. The dataset must include high-resolution images from different sources, including laboratory tests, field inspections, and simulation-based analyses, to provide a robust and varied training set that allows the AI model to generalize effectively to unseen data. These images are then annotated by domain experts who provide detailed descriptions, meticulously outlining key structural features, material properties, and potential defects such as cracks, voids, delamination, and corrosion patterns. Expert annotations are invaluable, as they establish a high-quality ground truth reference that ensures the AI-generated captions align with industry standards and convey meaningful insights regarding material conditions. The annotation process also involves multiple layers of review to validate the accuracy of descriptions and remove inconsistencies, ensuring that the dataset maintains a high level of reliability.
The first phase of this research focuses on the development of a Vision-Language Pretraining (VLP) model designed to generate high-quality, contextually accurate captions for NDE images. This phase consists of multiple stages, including data collection, preprocessing, model training, and performance evaluation. The data collection process involves compiling a comprehensive dataset of NDE images sourced from various imaging techniques, ensuring a diverse and representative dataset that captures a wide range of structural conditions, material defects, and subsurface anomalies. These images are then annotated by domain experts who provide detailed descriptions, meticulously outlining key structural features, material properties, and potential defects such as cracks, voids, delamination, and corrosion patterns. Expert annotations are crucial, as they establish a high-quality ground truth reference that ensures the AI-generated captions align with industry standards and interpret meaningful insights regarding material conditions. Once the dataset is ready, we move on to training the model. The setup combines a Convolutional Neural Network (ResNet-50) [9] to pull out visual features from NDE images and a transformer-based language model like GPT-2 to turn those features into meaningful captions. ResNet-50 extracts detailed structural info from the images, which is then fed into GPT-2, fine-tuned on NDE-specific data to keep the captions relevant and technical. We track performance using BLEU scores to see how well the captions match expert annotations, and domain experts also review the results to make sure the outputs are accurate and useful. As we go, we fine-tune things like learning rates and batch sizes to keep improving the model's performance. The end goal is to build a reliable system that can help engineers and inspectors quickly understand what NDE images are showing—saving time and improving accuracy in defect detection.

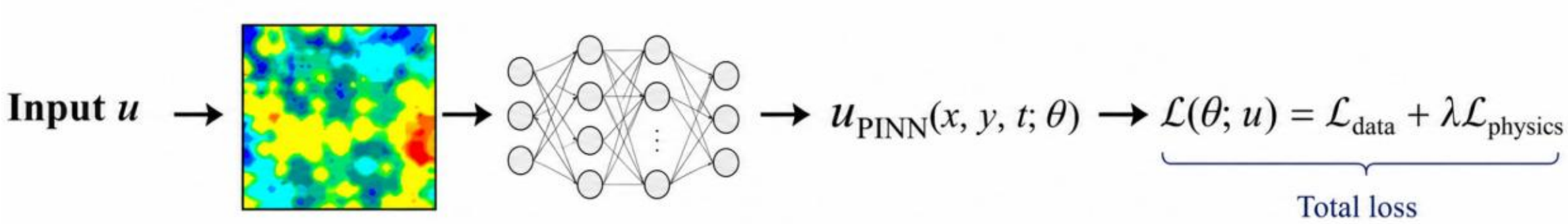

*Figure 1. A whole picture of phase 1 research*

Where

**Input u → Input Field → Neural Network → u_PINN(x, y, t; θ) → ℒ(θ; u) = ℒ_data + λℒ_physics**

The diagram represents a Physics-Informed Neural Network workflow in which input information is processed by a neural network to predict a physical response while training is constrained by both observed data and governing physical laws. The meaning of the characters and symbols is shown as follows:

| ***Character /Symbol*** | **Meaning** |
|---|---|
| *u* | The input or physical field variable. Depending on the application, it may represent displacement, temperature, electromagnetic-field amplitude, GPR signal intensity, or another physical quantity. |
| *u_PINN* | The physical response or solution predicted by the Physics-Informed Neural Network. |
| *PINN* | Physics-Informed Neural Network. |
| *x* | Spatial coordinate in the x-direction. |
| *y* | Spatial coordinate in the y-direction. |
| *t* | Time or temporal coordinate. |
| *θ (theta)* | The collection of trainable neural-network parameters, primarily weights and biases. |
| *;* | Separates the independent variables (x, y, t) from the model parameters θ in u_PINN(x, y, t; θ). |
| *ℒ* | The loss function used to train the neural network. |
| *ℒ(θ; u)* | The total loss evaluated using the network parameters θ and the available data or physical-field information u. |
| *ℒ_data* | Data loss. It measures the discrepancy between the neural-network predictions and the available measured, observed, or training data. |
| *$\mathscr{L}$_physics* | Physics loss. It measures how well the predicted solution satisfies the governing physical equations and associated constraints. |
| *λ (lambda)* | A weighting coefficient that controls the relative contribution of the physics loss to the total loss. |
| + | Indicates that the data loss and weighted physics loss are combined to construct the total training objective. |
| → | Indicates the direction of information flow through the PINN framework. |

*3.1.1. Interpretation of the Main Expressions*

**a. Input u**

The input represents the information supplied to the model. In a conventional PINN formulation, the independent variables (x, y, t) are often used directly as the network inputs, while u represents the physical quantity that the network predicts.

**b. u_PINN(x, y, t; θ)**

This expression represents the PINN-predicted solution. The coordinates x and y specify spatial location, t specifies time, and θ contains the trainable parameters learned during optimization. For a GPR application, u_PINN may represent a predicted electromagnetic field, GPR response, or another quantity associated with subsurface behavior.

**c. Total loss is shown as follows:**

$$\mathscr{L}(\theta; u) = \mathscr{L}_data + \lambda\mathscr{L}_physics$$

The total loss combines a data-driven term and a physics-based term. $\mathscr{L}$_data encourages agreement between the model predictions and observed data, whereas $\mathscr{L}$_physics encourages the predictions to satisfy the governing physical equations. The coefficient λ controls the balance between these two objectives.

**d.Recommended PINN Notation for a Research Paper**

If the diagram is intended to show conventional PINN architecture, a clearer formulation is to label the network inputs as (x, y, t) rather than Input u:

$$(x, y, t) \rightarrow PINN_\theta \rightarrow u_PINN(x, y, t) \rightarrow \mathscr{L}_data + \lambda\mathscr{L}_physics$$

This notation clearly distinguishes the independent input coordinates from the physical response predicted by the network and is generally easier for readers to interpret.

*3.2 Architecture*

The system we're building uses a hybrid deep learning setup that combines **ResNet-50** for pulling features from images and **GPT-2** for generating captions. ResNet-50 is a well-known convolutional neural network that does a great job of extracting detailed visual information from images [9]—like structural patterns, material properties, and signs of defects (Figure 3). Once it processes an image, it outputs a high-dimensional feature vector that captures all the important visual cues. To connect this with the language part of the system, we use a linear transformation layer that maps those image features into a format GPT-2 can understand. GPT-2, which is a powerful language model based on transformers, then takes over to generate natural language captions that describe what's in the image (Figure 4). It uses both its general language training and additional fine-tuning on NDE-specific data to keep the descriptions accurate and relevant. The GPT-2 model contains N Transformer decoder blocks, as shown in the left panel. Each decoder block (center panel) includes a multi-head masked attention layer, a multi-layer perceptron layer, normalization, and dropout layers. The residual connection allows the block to learn from the previous block's input. The multi-head masked attention layer (right panel) calculates attention scores using Q, K, and V vectors to capture sequential relationships in the input sequence.

Inside ResNet-50, the image goes through several layers that detect different levels of visual detail—from simple edges and textures to more complex shapes and structural issues like cracks or voids. These features are then processed through pooling layers, normalization, and fully connected layers to create a compact yet detailed 2048-dimensional feature vector. This vector gets translated into GPT-2's language space so the model can smoothly turn it into a clear, descriptive caption that makes interpreting the image much easier for engineers and inspectors. The whole process of image to caption steps is shown in **Error! Reference source not found.**.

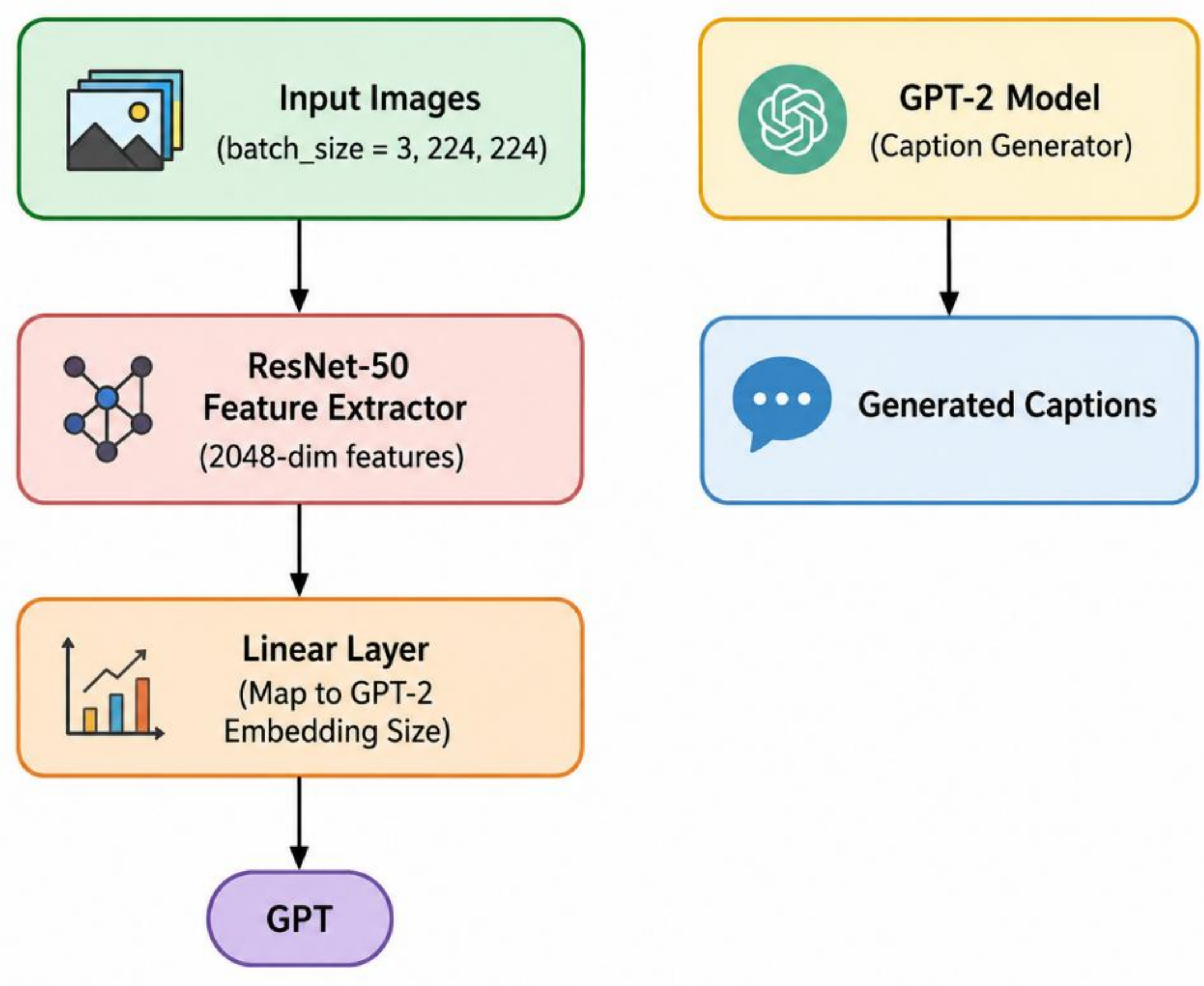

*Figure 2. The flowchart of whole process of image-to-caption steps.*

*3.2.1. The architecture of ResNet-50*

ResNet-50 is a deep convolutional neural network built using residual blocks, which help the model learn better by allowing it to skip layers and avoid problems like vanishing gradients [10]. The architecture is divided into five main parts: a Pre-net, four stages of residual layers, and a Post-net. It starts with the Pre-net (a few initial layers like convolution and batch norm) to process the raw input image. In Stage 1, it stacks 3 regular residual blocks (orange), each containing convolutional layers with skip connections, so the model can learn both simple and complex features without forgetting earlier ones. Stage 2 starts with a downsampling residual block (D. Res. Block, shown as a trapezoid) to reduce the feature map size, followed by 3 more residual blocks (blue). Stage 3 continues this pattern but gets deeper, with 1 downsampling block and 5 regular ones (green), allowing the model to learn even more complex and hierarchical features. Stage 4 also starts with a downsampling block and finishes with 2 regular ones (purple), wrapping up the deep feature extraction. Finally, the Post-net prepares the output features for the final classification layer (not shown here). The inset on the right shows the inner design of a residual block: an input goes through layers (like convolutions *f, g, h*), and then the output is added back to the original input using a skip connection—this helps the network learn "residual" information instead of trying to learn everything from scratch. Altogether, ResNet-50 has 50 layers and is known for being powerful and efficient for tasks like image classification, detection, and more (Figure 3).

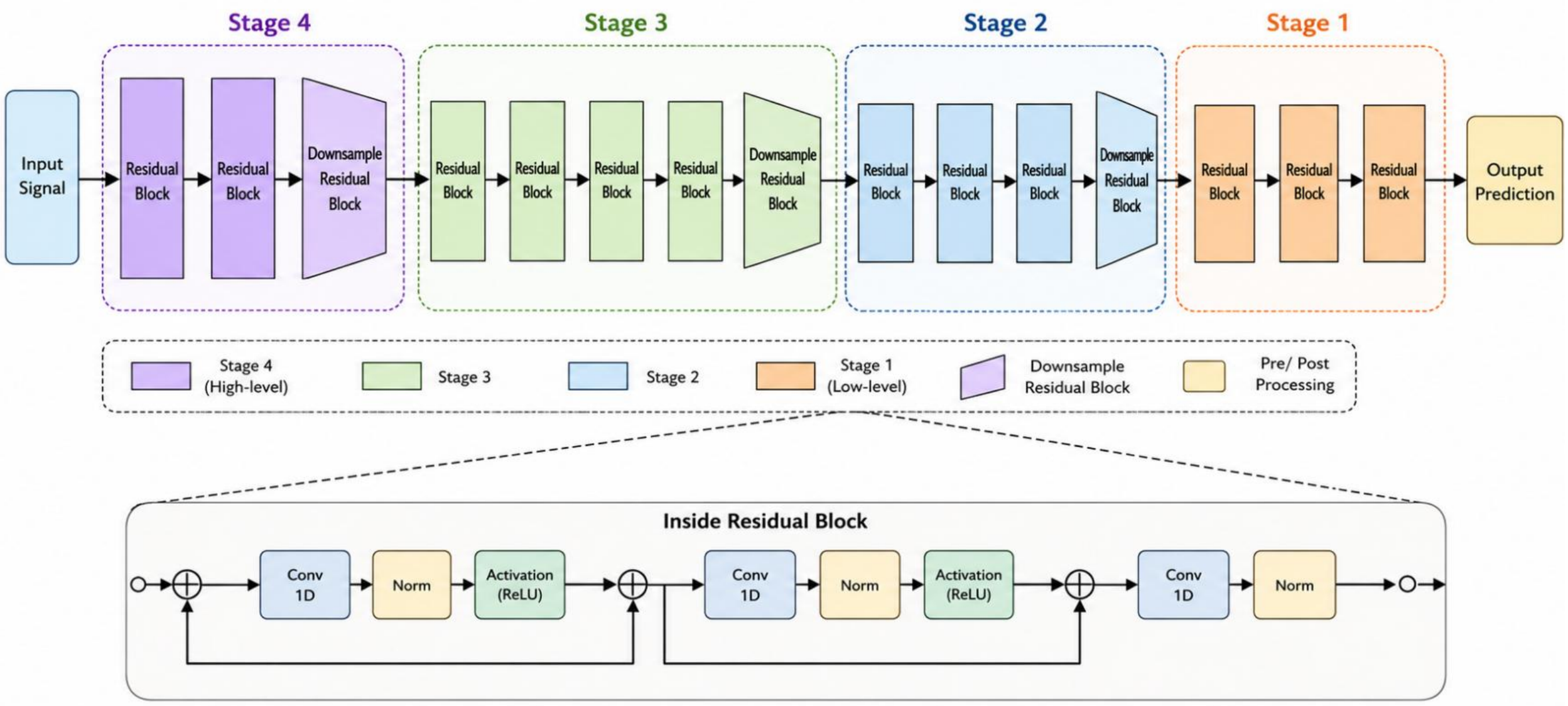

*Figure 3. The residual network architecture. Specifically, ResNet-50, which has 3, 4, 6, and 3 blocks in each stage, from input to output.*

*3.2.2. The architecture of GPT-2*

GPT-2 is a generative language model that predicts the next word in a sequence using a stack of transformer blocks [11]. The input consists of tokenized text, which is first passed through an embedding layer and combined with positional encodings to retain word order information. This processed input is fed into multiple identical transformer blocks, each consisting of two main components: a masked multi-head self-attention mechanism and a feedforward neural network. The attention mechanism allows the model to focus on relevant parts of the previous context while masking future tokens to prevent cheating during prediction. Each attention block uses query, key, and value projections, followed by *softmax* [12] and linear layers, and is computed across multiple heads in parallel to capture different types of relationships. After attention, the output goes through a feedforward network composed of two linear layers with a Gaussian Error Linear Unit (GELU) activation function [13]. Both sub-layers are wrapped with layer normalization, dropout for regularization, and residual connections to maintain gradient flow. The outputs of all transformer blocks are passed through a final layer normalization and a linear layer that projects the hidden states to vocabulary size, followed by a *softmax* layer that provides the probability distribution over possible next tokens. The model is trained to predict the next token in a sequence, and during inference, it generates text one token at a time by appending predictions to the input sequence (Figure 4).

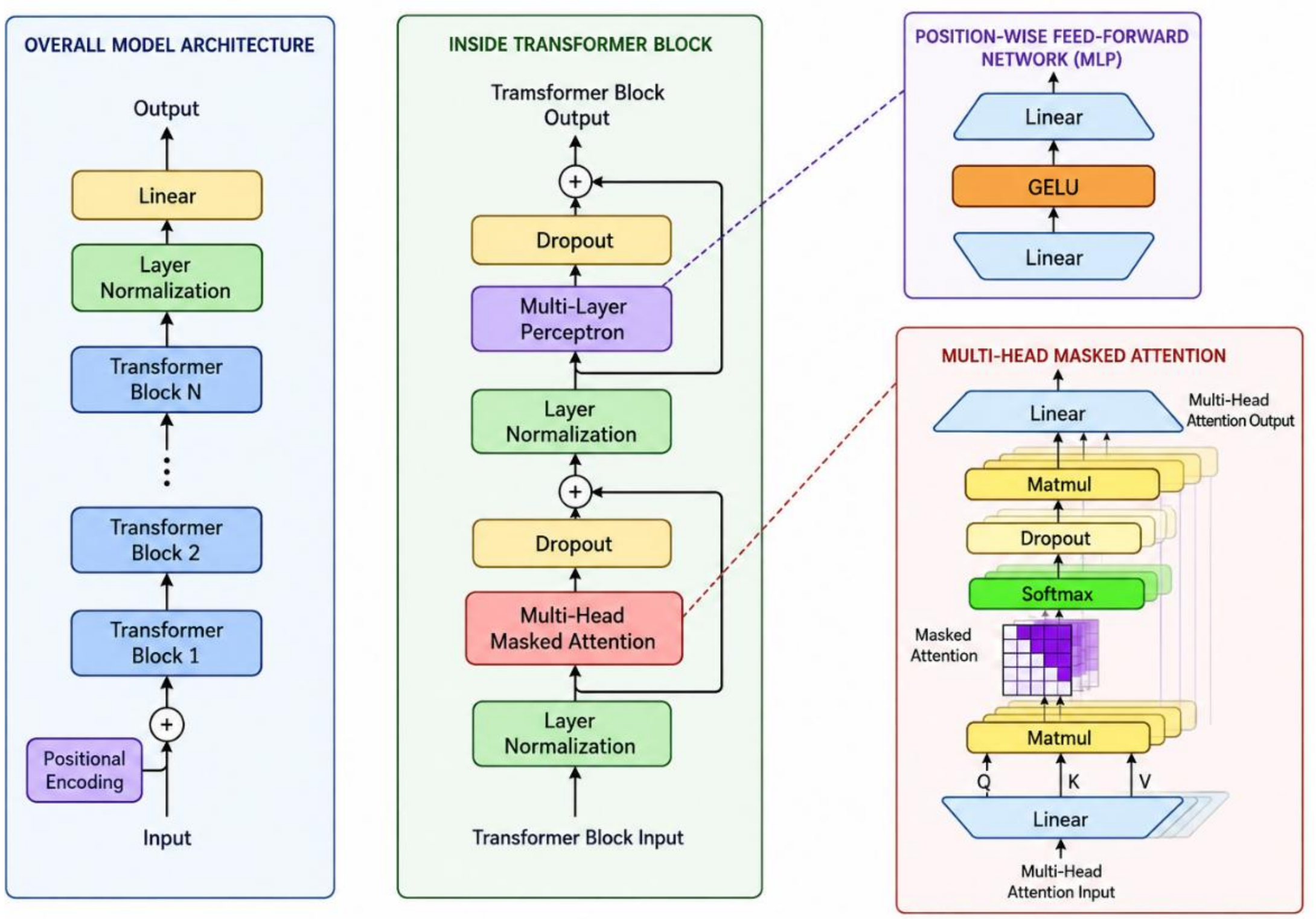

*Figure 4. GPT-2 model architecture.*

*3.3. Step by step data preparation*

To efficiently manage image-caption pairs for training an image captioning model, the images should be organized in a dedicated directory, such as *dataset/images/.* Alongside the images, a JSON file (e.g., captions.json) is created to store the corresponding captions. This JSON file follows a structured format, consisting of a list of dictionaries where each dictionary contains two keys: *image_path*, which specifies the relative path to the image, and *caption*, which holds the textual description of the image. This structured approach ensures seamless data loading and processing for training the model. To analyze the images, I utilized FHWA's InfoTechnology platform for interpretation (Figure 5). In the initial stage, captions were generated for each corresponding image. Following this, a JSON file was created to systematically store the images and their associated captions. An example of a JSON file is shown in Figure 6 . The training process incorporated Ground Penetrating Radar (GPR) and Impact Echo (IE) data from the BEAST dataset to enhance the model's learning capabilities. Moving forward, additional Non-Destructive Evaluation (NDE) data will be generated and integrated into the training process to further refine the model's performance and improve subsurface abnormality detection.

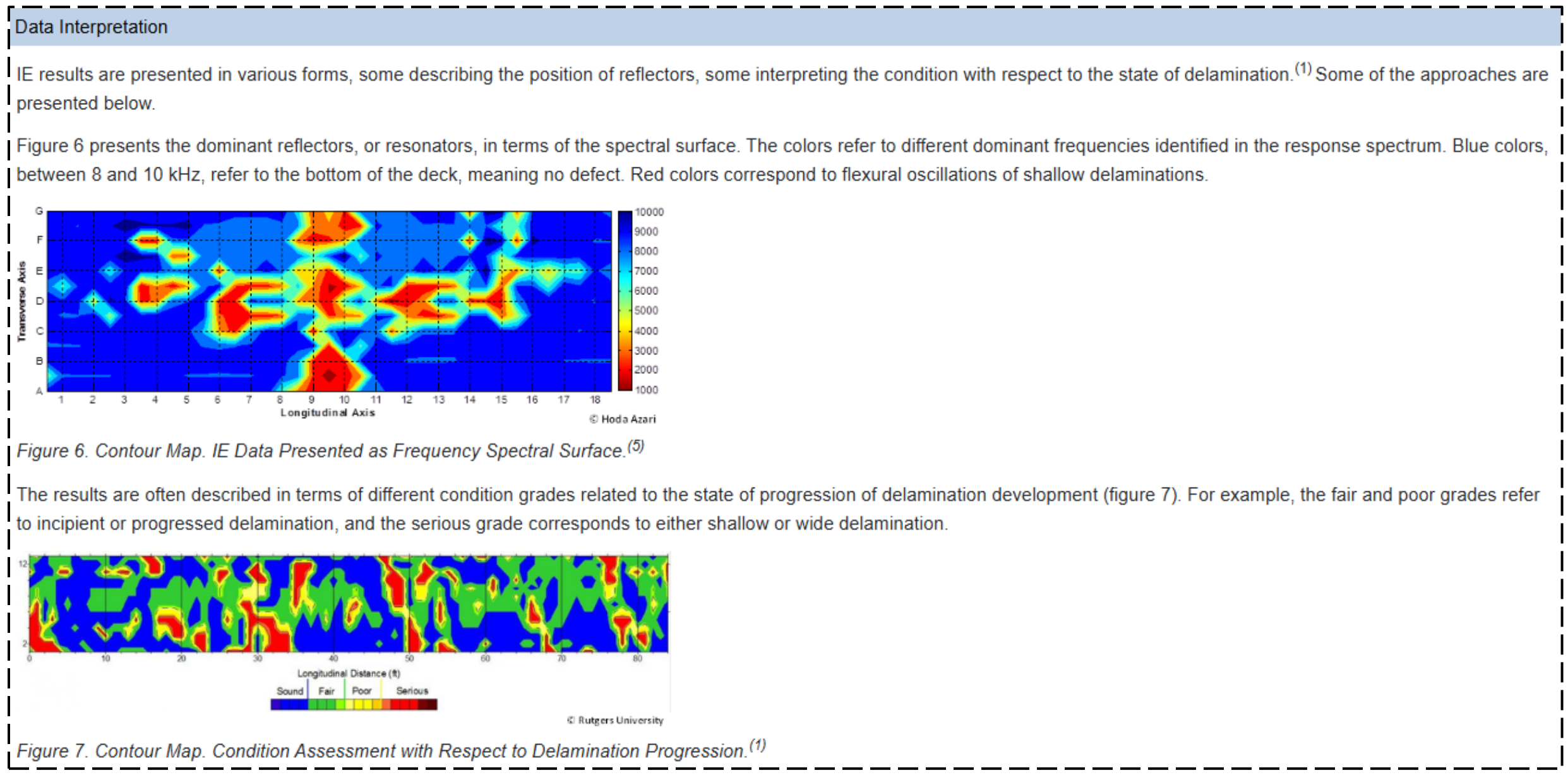

Data Interpretation

IE results are presented in various forms, some describing the position of reflectors, some interpreting the condition with respect to the state of delamination.[1] Some of the approaches are presented below.

Figure 6 presents the dominant reflectors, or resonators, in terms of the spectral surface. The colors refer to different dominant frequencies identified in the response spectrum. Blue colors, between 8 and 10 kHz, refer to the bottom of the deck, meaning no defect. Red colors correspond to flexural oscillations of shallow delaminations.

Figure 6. Contour Map. IE Data Presented as Frequency Spectral Surface.[5]

The results are often described in terms of different condition grades related to the state of progression of delamination development (figure 7). For example, the fair and poor grades refer to incipient or progressed delamination, and the serious grade corresponds to either shallow or wide delamination.

Figure 7. Contour Map. Condition Assessment with Respect to Delamination Progression.[1]

*Figure 5. InfoTechnology website to interpret images.*

```
[
    {
        "image_path": "C:/Users/m.shafiei.dizaji.ctr/Desktop/dataset/images/image1.jpg",
        "caption": "Impact Echo (IE) image illustrating subsurface response variations, with color intensities corresponding to response values. Low-r
    },
    {
        "image_path": "C:/Users/m.shafiei.dizaji.ctr/Desktop/dataset/images/image2.jpg",
        "caption": "Impact Echo image depicting variations in subsurface response, with color intensities quantified by response values. Low-intensity
    },
    {
        "image_path": "C:/Users/m.shafiei.dizaji.ctr/Desktop/dataset/images/image3.jpg",
```

*Figure 6. JSON file example*

### *3.4. Experiment Setup: Bridge Evaluation and Accelerated Structural Testing (BEAST)*

BEAST is the first full-scale bridge testing facility located in Piscataway, New Jersey (Figure 7). The BEAST facility is built and commissioned by the Center for Advanced Infrastructure and Transportation at Rutgers, The State University of New Jersey. The BEAST facility is the first facility nationwide capable of applying controlled and accelerated live load, environmental, and maintenance demands on full-scale bridge superstructures. The specimen is a multi-girder steel composite bridge (30 by 50 ft) with an 8-inch bare concrete deck and black rebar reinforcement. It will be subjected to rapid-cycling environmental changes and extreme traffic loading to speed up deterioration, as much as 30 times, in order to simulate 15-20 years of wear-and-tear in just a few months. The deck is supported by four I-beams as the main girders with one fixed and one open joint. At its initial design, the specimen is supposed to be exposed to over 8 million cycles of live loading (60 kips), 400 freeze-thaw and hot-dry cycles, as well as the application of deicing agents (6% brine solution) to simulate common winter maintenance practices. The primary target is to validate performance models by measuring stress and deterioration caused by live, environmental, and maintenance loading in an extremely compressed time frame. To that end, the BEAST experiment aims to utilize accelerated testing of the full-scale bridge deck and superstructure systems subjected to cyclic moving wheel loads and freeze-thaw environmental conditions. It is primarily envisioned that BEAST experiments will complement both field observations and material-level tests and fill an important gap in our current understanding of bridge performance and deterioration. The BEAST database encompasses data collected through five Nondestructive Evaluation (NDE) technologies including Impact Echo (IE), Ground

Penetrating Radar (GPR), Ultrasonic Surface Waves (USW), Electrical Resistivity (ER), and Half-Cell Potential (HCP). This comprehensive data collection was primarily conducted on the BEAST bridge specimen, focusing on the concrete bridge deck. Data use in this study are the processed BEAST data that can be found on the https://infobridge.fhwa.dot.gov/ website. In this work only IE and GPR data is used to train the model. In the Table 1 number of samples to train, validation, and testing is shown.

*Figure 7 .The overview of the BEAST facility.*

*Table 1. Number of samples to train, validation, and testing*

| | | Number of samples | | |
|---|---|---|---|---|
| **Tasks** | **Data** | **Train** | **Val** | **Test** |
| **VLP** | Image-caption pairs | 550 | 35 | 10 |

*3.5. Step by step training process*

In this setup, the *AdamW* optimizer [14] from the Transformers library is employed to fine-tune the GPT-2 model, enabling it to generate more accurate captions based on the given images and text data. The training loop starts by tokenizing the input texts for each epoch and batch. These texts are then repeated to match the batch size of the images being processed, ensuring that the text and image data are aligned in terms of size. The target texts, which consist of both the initial prompt and the actual captions, are tokenized to create labels for the model. These labels are essential for guiding the model during training, ensuring it learns the correct relationships between the provided images and the generated captions. Each batch begins by zeroing the gradients, clearing any previous accumulated gradients from prior training steps. This is crucial for ensuring the accuracy of the backpropagation process. Once the gradients are reset, the model performs a forward pass, during which the input data is passed through the model to compute the loss, representing the difference between the predicted captions and the target captions. After computing the loss, the gradients are calculated via backpropagation, and the optimizer adjusts the model's parameters to minimize the loss.

This process is repeated for each batch and each epoch, gradually improving the model's performance. At the end of the training, the model is saved for later use, ensuring that the fine-tuned version is preserved for generating captions on new data. Hyperparameters such as ***num_epochs*** and ***learning_rate*** are adjusted based on the specific dataset and available computational resources. Fine-tuning these hyperparameters can have a significant impact on the training process, allowing the model to generate more accurate and relevant captions. The flexibility to tweak these values enables optimization based on the unique characteristics of the dataset, ensuring that the model can effectively capture the underlying relationships between images and captions for improved performance.

### *3.3 Performance Evaluation*

To see how well the VLP model is actually performing, we use the **BLEU score** [15]—a common metric for checking how close the AI-generated captions are to the expert-written ones. It works by comparing small chunks of words (called n-grams) from both the model's output and the reference captions. BLEU also includes a penalty for captions that are too short, to make sure the model doesn't just generate vague or overly simple sentences. The results so far show that the model does a good job at capturing the overall structure and identifying common defects. But when it comes to generating longer, more detailed captions, it still has some trouble. One of the main issues is that the model tends to simplify its descriptions, which makes them less accurate compared to what human experts would write. This usually happens because long-text generation is tricky—transformer models like GPT-2 sometimes struggle to keep things coherent and specific over longer outputs. To fix this, we use VQA model to make captions shorter and more specific.

#### *3.3.1. Bilingual Evaluation Understudy (BLEU) metric*

We first evaluated the accuracy using the BLEU score, proposed by Papineni et al. (2002) [16]. The score is derived as follows:

$$\mathrm{BLEU} = \mathrm{BP} \cdot \exp\left(\sum_{n=1}^{N} w_n \log p_n\right)$$

$$\mathrm{BP} = \begin{cases} 1 & \text{if } c > r \\ e^{(1-r/c)} & \text{if } c \leq r \end{cases}$$

where r is the length of the correct answer, c is the length of the output sentence, and BP is the penalty given if the output sentence is shorter than the correct answer. Further, *pn* is defined as the modified precision score as shown in Equation (XX).

$$p_n = \frac{\sum \begin{array}{l}\text{Number of n} - \text{grams that matched} \\ \text{in output text and correct answer}\end{array}}{\sum \text{Number of n} - \text{grams in the output text}}$$

The n-gram is a sequence of *n* adjacent words, where N is the largest n. In this study, we set N = 1 to 4. Here, *wn* is the weight, which is often calculated as *wn* = 1/N and was used in this study as well.

### *3.4. Results using BLEU matrix for VLP model*

The evaluation of the Vision-Language Pretrained (VLP) model using the BLEU metric provides valuable insight into its ability to interpret and describe Impact Echo (IE) images commonly used in Non-Destructive Evaluation (NDE) for subsurface assessments. The training loss curve, depicted in the graph, shows a steep decline within the first 50 epochs, followed by a plateau phase with minor fluctuations, indicating that the model successfully converged to a relatively stable state with minimal loss around epoch 250 (Figure 8). Despite the promising convergence behavior, the BLEU scores (Table 2)—BLEU-1 (0.37), BLEU-2 (0.29), BLEU-3 (0.25), and BLEU-4 (0.22)—suggest that the model's language generation capability remains moderate and leaves room for improvement, particularly in generating higher-order n-gram matches. The comparison between the ground truth (GT) and the predicted model (PM) descriptions highlights several qualitative differences. The **GT** caption provides a comprehensive and technical explanation of the subsurface conditions depicted in the image, incorporating specific response value ranges (e.g., 1000–5000 for low-intensity and 9000–10000 for high-intensity regions), depth interpretations, material consistency implications, and suggestions for structural evaluation. In contrast, the PM caption tends to generalize the content, capturing only the broad concepts—such as the presence of surface defects or intact materials—without offering detailed numerical interpretations or nuanced language structure (Figure 9).

*Table 2. Evaluation of the trained model using BLEU Matrix*

| Matrix | BLEU-1 | BLEU-2 | BLEU-3 | BLEU-4 |
|---|---|---|---|---|
| **Accuracy** | 0.37 | 0.29 | 0.25 | 0.22 |

From what we've seen, the relatively low BLEU scores aren't too surprising—there are a few common hurdles that tend to show up in vision-language tasks, especially when you're working in specialized fields like NDE. For starters, the language used in engineering reports is full of technical terms and structured phrasing that most general-purpose language models just aren't trained on. So even if the AI-generated caption gets the meaning right, the wording might be off enough to hurt the BLEU score. Another thing to keep in mind is that BLEU focuses a lot on exact word matches and short phrase overlaps (n-grams). That means if the model paraphrases something—even if it's totally accurate—it can still get penalized. Plus, our training data is pretty limited and highly specific, so the model doesn't have a ton of variety to learn from. This makes it harder for it to generalize when it sees new image patterns or different ways of describing the same defect. And finally, BLEU is all about precision—it doesn't really consider how relevant or contextually accurate the caption is, which is super important in fields like NDE where certain words carry a lot of weight. To boost performance, we're thinking about fine-tuning the language model on a bigger, more targeted dataset that includes expert-written NDE captions. That would help the model get more familiar with the right vocabulary and writing style. We're also exploring ways to expand the dataset using simulated or synthetic IE images, paired with high-quality captions, to give the model more exposure to different visual scenarios. On top of that, switching to other evaluation metrics like **METEOR** or **BERTScore** [17]—which are better at capturing the meaning behind the words—might give us a clearer picture of how well the model is actually doing.

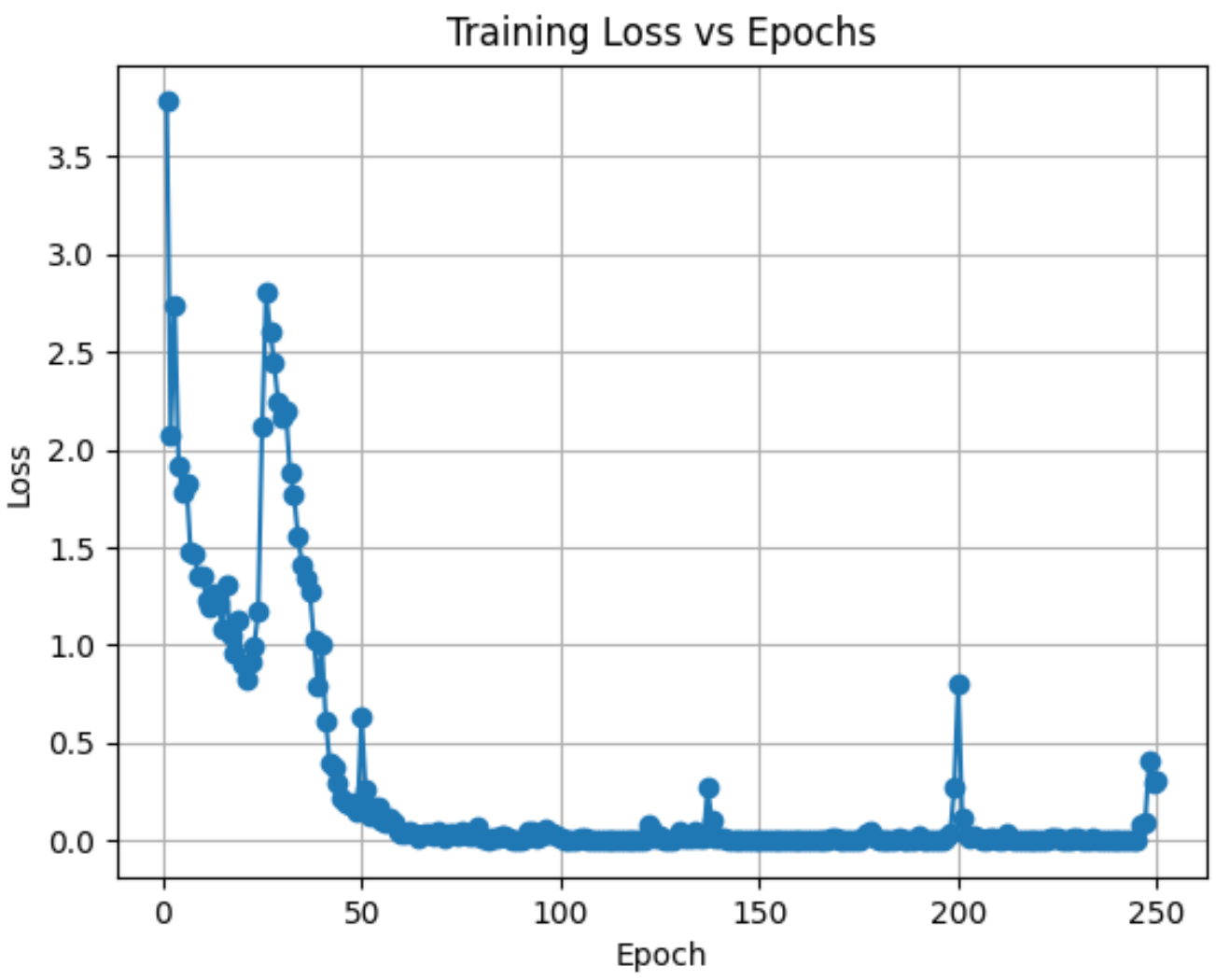

*Figure 8. Training loss vs Epochs*

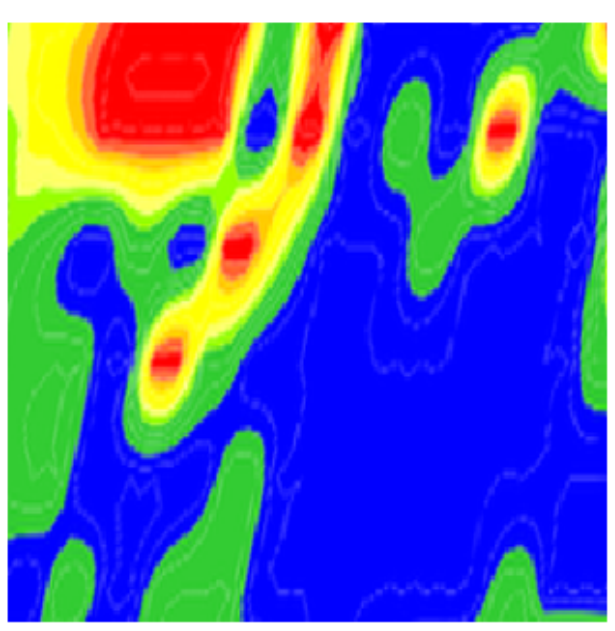

**GT**: *"Impact Echo (IE) image illustrating subsurface response variations, with color intensities corresponding to response values. Low-response areas (in red to yellow, values around 1000–5000) are concentrated in the upper-left section, suggesting potential surface-level defects, such as voids or delaminations, indicating reduced material integrity in these regions. Intermediate-response zones (in green, values around 5000–7000) are observed along the edges and near the lower-left, representing areas with partial material consistency. High-response regions (in blue, values around 9000–10000) dominate the lower portion of the image, indicating intact material with strong, uniform properties at greater depths. The localized concentration of low-response areas in the upper section suggests possible structural weaknesses that may need further investigation."*

**PM**: *Impact Echo image shows subsurface variations. Low-response suggests surface defects, medium indicates partial consistency, and high signals strong material. Clusters of low response may indicate structural issues*

*Figure 9. Testing the trained model for caption generation*

## 4. Phase 2: Visual Question Answering (VQA) Model

Visual question answering is a task that requires answering questions about a scene based on an input image and natural language text. With the advent of deep learning, significant progress has been made in fundamental tasks such as image classification, motivating researchers to explore higher-level tasks that involve both vision and language, such as VQA. Early studies on VQA focused on extracting good features from the input image and fusing the two modalities to predict the correct answer. To extract features from the input question, existing natural language processing techniques have been adapted. The training process for the Visual Question Answering (VQA) model was conducted in two structured stages to enhance the model's multimodal reasoning capabilities. In the first stage, the model was trained on large-scale image–text pairs, which helped establish a foundational understanding of the relationships between visual content and corresponding textual descriptions. This phase, aligned with Phase 1 of the training strategy, allowed the model to learn generic visual-language representations such as spatial relationships, and contextual language associations. In the second stage, the model was fine-tuned on a domain-specific VQA dataset, carefully generated to include questions and answers explicitly grounded in visual content. The fine-tuning stage was critical for adapting the general-purpose vision-language model to the structured reasoning and answer-generation demands of the VQA task. Visual Question Answering (VQA) models are designed to interpret and answer questions about images by integrating techniques from computer vision and natural language processing. A typical VQA model architecture involves several key components:

1. **Image Feature Extraction**: Utilizes Convolutional Neural Networks (CNNs) to process the input image and extract meaningful visual features.
2. **Question Processing**: Employs Transformer-based models to encode the input question into a feature representation.
3. **Feature Fusion**: Combines the visual and textual features through attention mechanisms, to focus on relevant parts of the image in the context of the question.
4. **Answer Prediction**: Generates a response based on the fused features, typically using a classifier to select the most appropriate answer.

*4.1. Methodology*

Let us assume a set $D = \{X_i, q_i, y_i\}$ of a NDE image, a question, and the corresponding answer. The goal is to teach the model to generate the correct answer $yi$ to the question $qi$ about the given NDE image $Xi$. Figure 10 shows the overall framework of our NDE VQA model, which is composed of an image encoder, a question encoder, and an answer decoder.

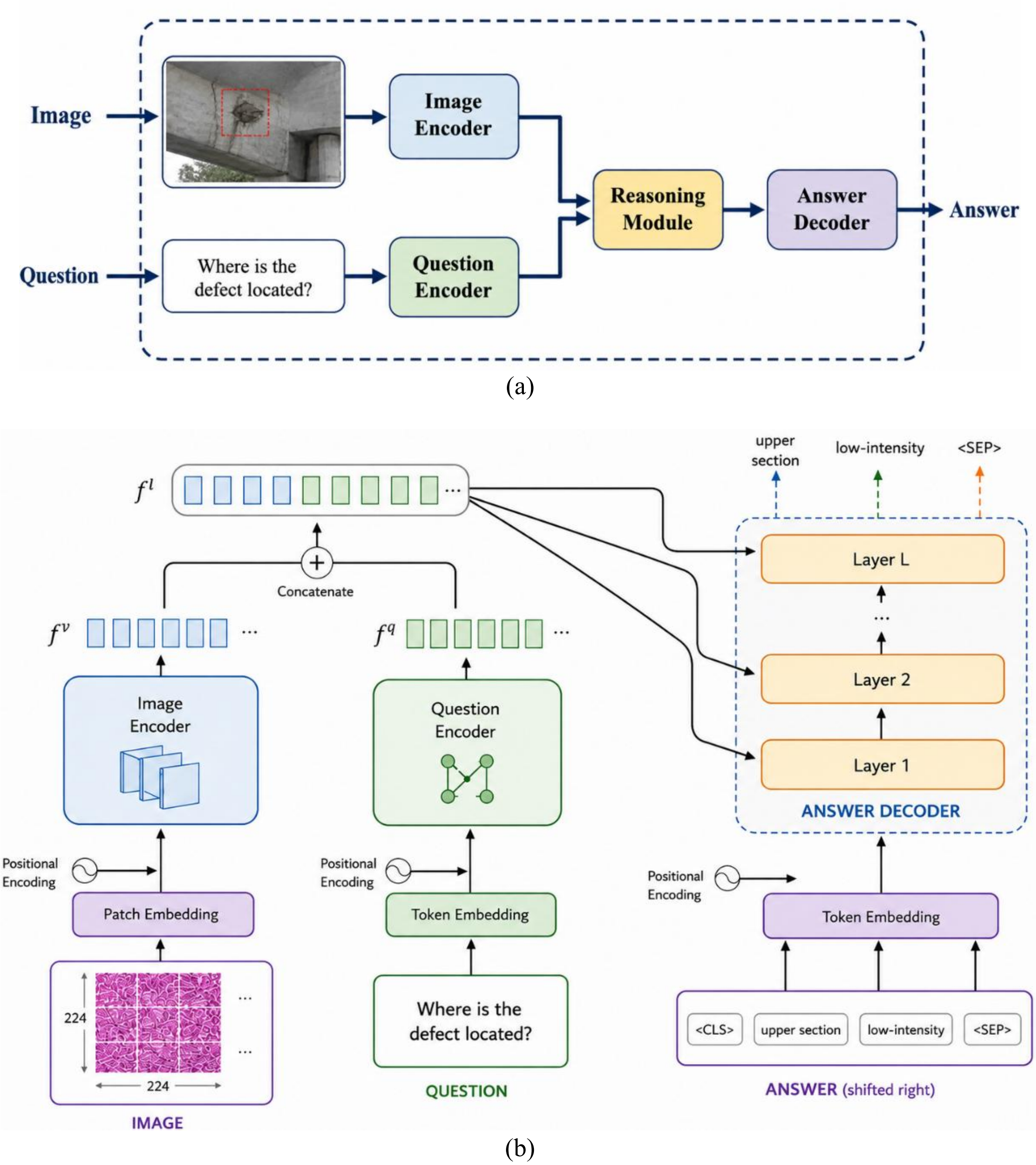

*Figure 10. (a) General architecture of the proposed medical VQA model. (b) The details of the VQA model.*

The proposed model uses separate encoders for each input modality, followed by a shared decoder. A transformer-based visual encoder processes the input NDE image to extract visual features, while a language encoder converts the input question into a corresponding language representation. Once both modalities are encoded, their feature vectors are combined to form a unified multimodal representation. This combined representation is then passed through multiple layers of the decoder to generate the final answer. The following section provides a detailed overview of the model architecture.

*4.1.1. Image Encoder*

Figure 2 shows a closer look at how the image encoder is built. The NDE image $X_i$ goes into the encoder to create its visual representation. First, the image is resized to 224×224×3 pixels. After that, it's split into 49 separate patches, each one 32×32 pixels in size and non-overlapping. These patches are then flattened into one-dimensional vectors and passed through an embedding layer that transforms them into vectors of size 768, so they match the input size expected by the encoder. Finally, positional encodings are added to these patch embeddings to preserve spatial information before feeding them into the image encoder.

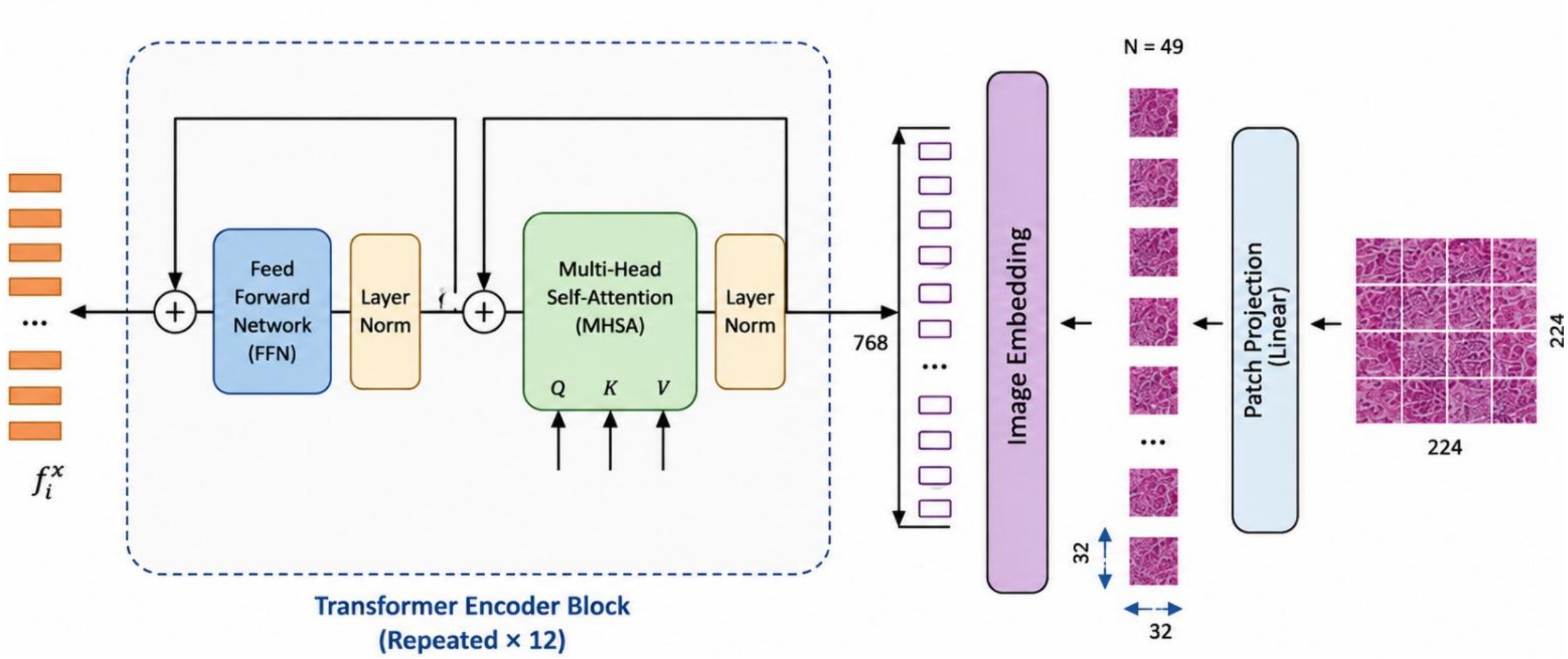

*Figure 11. The architecture of the image encoder.*

The image encoder used here is the ViT32 model, a version of the Vision Transformer introduced in [18]. This model usually has 12 identical layers, and each one includes a multi-head self-attention (MSA) block and a feed-forward network (FFN) that works together to extract visual features from the input. Before each of these blocks, there's a normalization step, and the model also uses residual connections to help pass information smoothly from one layer to the next block [19]. The Multi-Head Self-Attention (MSA) [6] in the encoder uses a self-attention mechanism to figure out how different patches of the NDE image are related to each other. To do this, the embedded image is passed through three separate linear layers to create three key components: the query (Q), key (K), and value (V) matrices. The model then calculates the dot product between Q and K to measure the similarity between patches. This result is scaled by dividing it by the square root of the dimension of K, and then it goes through a SoftMax function to generate attention weights. Finally, these weights are used to scale the V matrix, producing a weighted combination that reflects the most relevant parts of the image. This process is summarized in the following formula:

$$Attention = (Q, K, V) = Softmax\left(\frac{QK^T}{\sqrt{d_K}}\right).V$$

In the MSA block, several self-attention heads run in parallel, each computing its own version of scaled dot-product attention. The outputs from all these heads are then combined (concatenated) and passed into the feed-forward network (FFN). The FFN is made up of two fully connected layers with a GELU activation function in between. Once the image has been processed through the encoder, its representation is projected down to a 512-dimensional vector, so it lines up with the question's representation. In the end, the image is represented as a 49 × 512 matrix, where 49 is the number of patches.

*4.1.2. Question Encoder*

The question encoder uses a BERT-like architecture [20] to generate the question's textual features. Similar to the image encoder, the question encoder consists of a stack of 12 identical layers. As shown in Figure 12, the first step in encoding the question is tokenization, in which the question is tokenized as a sequence of word tokens. A learnable positional embedding is added to the sequence to provide information about the order of each word. The final representation is generated by feeding the initial representation through the 12 layers of the encoder. Analogously to the image encoder, the question encoder employs the MSA block to capture dependencies within the question tokens. The model also uses normalization layers and skips connections, but unlike the image encoder, the normalization layers come after the MSA and FNN. The output of the question encoder is the question feature representation of size 77 × 512. This representation holds information about the semantics of the question and the relationships between words.

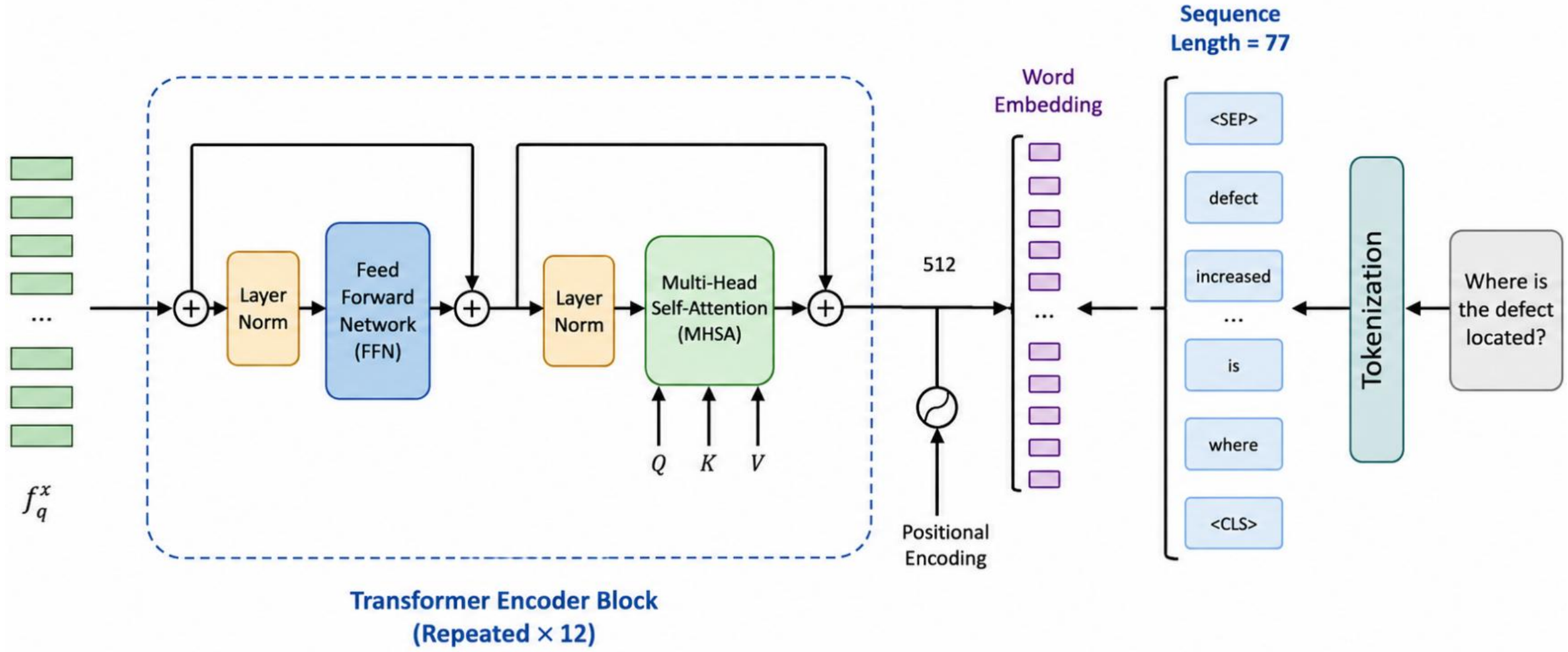

*Figure 12. The architecture of the text encoder*

*4.1.3. Answer Decoder*

The decoder works like a generative model that builds the answer one word at a time. It does this in an autoregressive way—meaning, every time it predicts a word, it adds that word to the input and uses the updated sequence to predict the next one. The decoder has two identical layers, and Figure 13 gives a look inside one of them. Each decoder layer includes components similar to the encoder—like Multi-Head Self-Attention (MSA) and a Feed-Forward Network (FFN)—but with a couple of key differences. One big change is that the decoder uses masked self-attention, which means it only looks at the previous words in the sequence, not the ones that come after. This allows the model to predict the next word based only on what it has already seen. Unlike self-attention, which uses *Q, K, and V* from the same input, cross-attention takes the query (*Q*) from the multi-modal input (combined image and question info), and the key (*K*) and value *(V)* from the current answer sequence. This setup helps the model understand how the different pieces of information relate, which is super useful for answering complex visual questions.

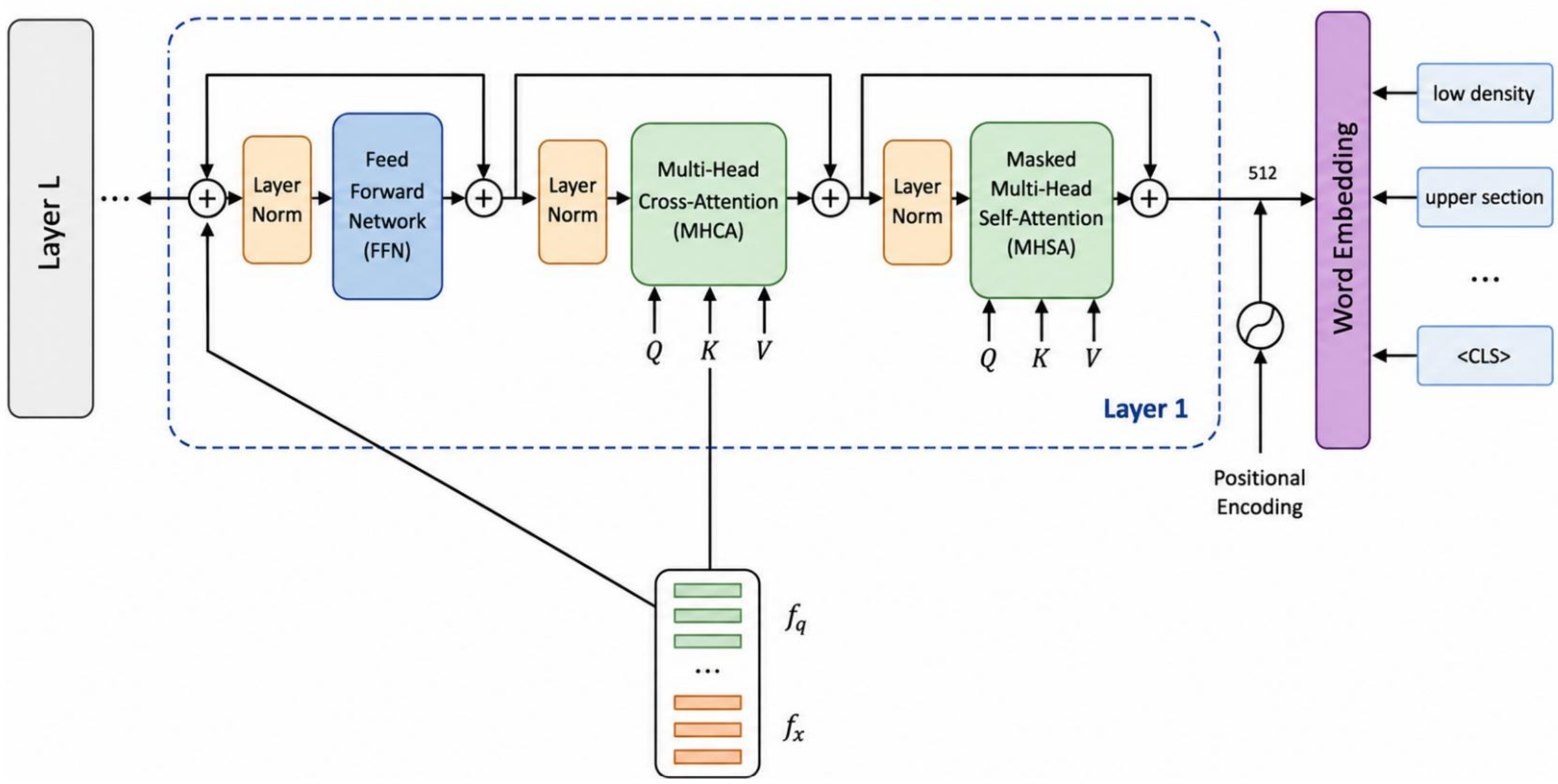

*Figure 13. The architecture of the answer decoder.*

*4.1.4. Evaluation Measures*

BiLingual Evaluation Understudy [16] (*BLEU*) is automatic evaluation metric to measure the similarity of predicted answers and ground-truth by matching n-grams, as expressed below:

$$BLEU = BP.\, e^{\sum_{n=1}^{N} w_n log_e P_n}$$

where $BP$ is the brevity penalty to penalize short answers, $wn$ is the weight between 0 and 1 for $log_e P_n$ and $\sum Nn=1wn=1$, $P_n$ is the geometric average of the modified n-gram precision, and $N$ is the maximum length of n-grams. N-grams here are up to 4.

*4.2. Dataset Preparation*

The second phase of this research focuses on developing a Visual Question Answering (VQA) model, which enables users to interact with NDE data more dynamically by posing questions about specific aspects of the image and receiving intelligent, context-aware responses. The dataset preparation process for VQA builds upon the annotated image-caption pairs from Phase 1, transforming them into structured question-answer formats. This involves curating a diverse set of domain-specific questions that address critical attributes of NDE images, including defect types, locations, dimensions, severity levels, and risk assessments. Expert-verified answers are then integrated into the dataset to ensure consistency and accuracy, creating a robust foundation for training an AI model capable of answering complex queries related to structural conditions. Table 3 shows the number of samples to train, validation, and testing.

*Table 3. Number of samples to train, validation, and testing*

| | | Number of samples | | |
|---|---|---|---|---|
| **Tasks** | **Data** | **Train** | **Val** | **Test** |
| **VQA** | Image question answer tuples | 550 | 35 | 10 |

*4.3. Evaluation*

The VQA model's performance is evaluated using the BLEU score [16], which measures how closely the model's answers match the correct ones by comparing small word groups, or n-grams. The results show that the VQA model performs better than the captioning model, mainly because its answers are shorter and more to the point, which helps avoid errors from being too wordy. On top of that, a closer look at the results shows that the model does a great job spotting important structural defects, correctly identifying their features, and giving solid risk assessments. By letting users ask specific questions about NDE images, the VQA model makes the data much easier to understand and use, which is a big plus for infrastructure diagnostics and structural health monitoring.

*4.4. Example Questions & Model Responses*

The following are examples of questions posed to the VQA model, along with corresponding ground-truth (GT) and predicted model (PM) responses:

1. **What type of defect is observed in the image?**
    - *GT: "Surface-level defect, including delamination."*
    - *PM: "Possible surface defect, delamination."*
2. **Where is the defect located?**
    - *GT: "Upper section with low-intensity response values."*
    - *PM: "Defect near the upper section."*
3. **What is the defect's severity?**
    - *GT: "Potential structural weaknesses requiring further inspection."*
    - *PM: "Possible structural weakness."*
4. **What are the color-coded regions representing in this image?**
    - *GT: "The red and dark red regions indicate severe material deterioration, while blue regions represent intact structural integrity."*
    - *PM: "Red shows deterioration, blue indicates intact material."*

*4.5. Results & Discussion*

*4.5.1. Impact Echo images*

The VQA model's performance was measured using the BLEU metric, and the results show that its answers closely match the correct, expert-provided ones. As seen in Table 4, the model scored 0.68 for BLEU-1, 0.65 for BLEU-2, 0.61 for BLEU-3, and 0.51 for BLEU-4—pretty solid numbers that show it's good at giving accurate and well-worded responses. For this test, the input was an Impact Echo image paired with a domain-specific question: “What type of defect is observed in the image?” The model correctly picked up on the low-intensity zones (the red-to-yellow areas with values between 1000 and 5000) near the surface, and it responded with an answer pointing to surface-level delamination (Figure 14). While its response ("PM") was a bit more concise than the expert-labeled answer, it still captured the key info needed to identify the defect. These BLEU scores, especially when compared to typical captioning tasks—suggest that fine-tuning the model on a specialized VQA dataset really helped sharpen its reasoning and response generation. Overall, the model shows it can understand detailed visual patterns and answer technical questions with accuracy, which is a big win for real-world use in non-destructive testing and structural evaluations.

*Table 4. Evaluation of the trained model using BLEU Matrix*

| Matrix | BLEU-1 | BLEU-2 | BLEU-3 | BLEU-4 |
| --- | --- | --- | --- | --- |
| Model | 0.68 | 0.65 | 0.61 | 0.51 |

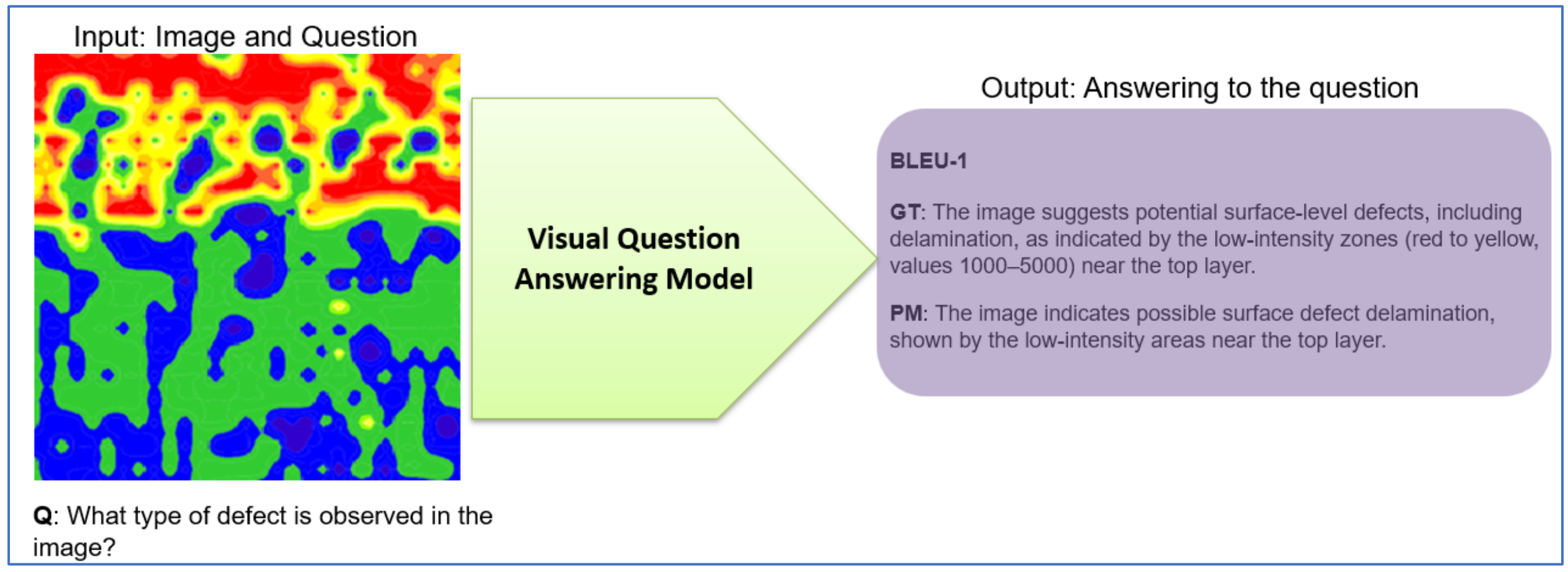

*Figure 14. The input for VQA model is an Impact Echo image paired with a domain-specific question: “What type of defect is observed in the image?”*

Continuing the VQA model evaluation using the BLEU metric, a second question was asked to see how well the model could pinpoint where a defect is located in the material just by looking at the image. The same Impact Echo image was used, but this time the question was: “Where is the defect located within the material or component?” (Figure 15). The model’s answer was compared to expert-labeled ground truth, and again, the BLEU scores were strong—0.68 for BLEU-1, 0.65 for BLEU-2, 0.61 for BLEU-3, and 0.51 for BLEU-4. The expert answer pointed out that the defects were mainly near the top layer in low-intensity areas (values between 1000 and 5000), which suggests possible delamination or surface flaws. The model’s response matched up well, noting that the defects were near the surface in those same low-intensity zones. Even though its answer was a bit shorter, it still captured the key technical points. This shows that the model is reliable across different types of VQA tasks, and that fine-tuning it on domain-specific questions really helped it learn how to reason about visual data. Along with the earlier example, this result proves that the VQA model can handle both identifying what kind of defect is present and figuring out where it is—making it a really useful tool for automated inspections in NDE applications.

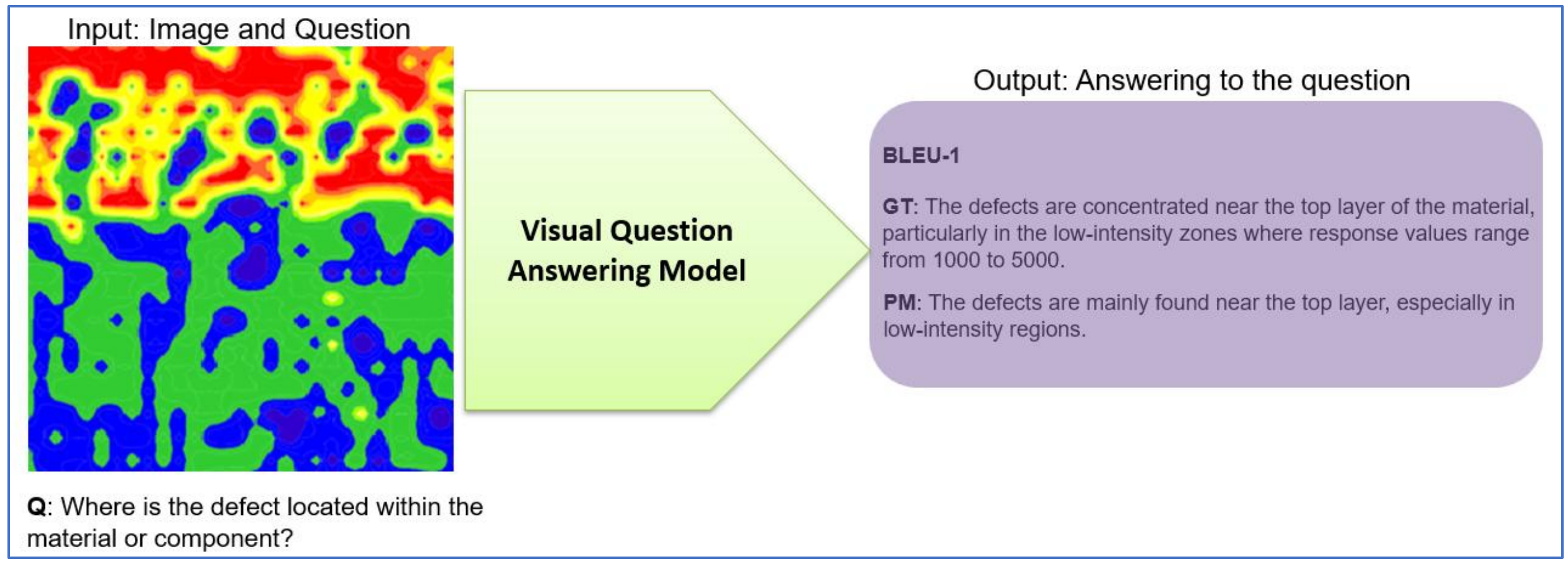

*Figure 15. The input for VQA model is an Impact Echo image paired with a domain-specific question: "Where is the defect located within the material or component?"*

The third VQA model evaluation looked at how well it could figure out the shape of a defect in an Impact Echo image. The question was: "What is the shape of the defect—linear, circular, irregular, or something else?" (Figure 16). This task went a step further than just spotting or identifying defects; it required the model to interpret the likely geometric form of what it was seeing. The BLEU scores stayed consistent with previous tests—0.68 for BLEU-1, 0.65 for BLEU-2, 0.61 for BLEU-3, and 0.51 for BLEU-4—showing that the model kept up its solid performance across different types of questions. The expert answer noted that, while the shape isn't clearly labeled in the image, defects like voids and delamination often show up in irregular or layered patterns. The model's answer was along the same lines, suggesting delamination usually appears in an irregular form—though it said it more briefly. This example points to a tricky part of VQA for NDE: questions that require inference or domain understanding (like guessing shape) can be abstract and not directly visible. Even so, the model gave a reasonable and accurate response that matched expert insights. Taken together with the previous results on defect type and location, the consistent BLEU scores show the model has solid potential for helping automate complex NDE image analysis.

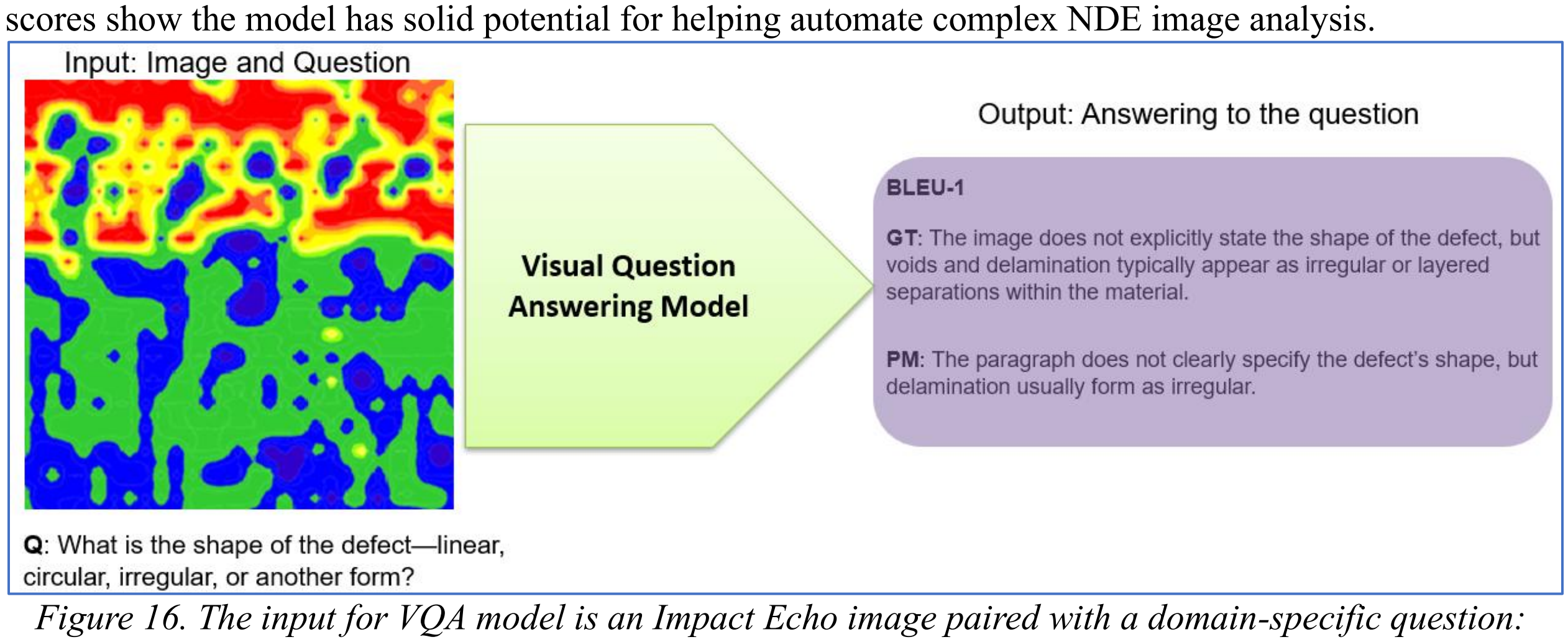

*Figure 16. The input for VQA model is an Impact Echo image paired with a domain-specific question: "What is the shape of the defect—linear, circular, irregular, or something else?"*

In the fourth VQA test, the model was asked to judge how **serious** the defect might be by answering: "How severe is the defect in terms of its potential impact on the material or component?" (Figure 17). This kind of question pushes the model beyond just describing or locating defects—it has to make a judgment call based on what the image shows and what that means for the structure's health. Once again, the model kept up its solid BLEU scores—0.68 for BLEU-1, 0.65 for BLEU-2, 0.61 for BLEU-3, and 0.51 for BLEU-4—showing its responses stayed consistent and well-formed. The expert answer explained that the cluster of low-intensity values near the surface could point to structural issues that might need closer inspection. The model picked up on the same idea, correctly flagging those surface-level, low-intensity zones as possible indicators of damage—though it explained things a bit more briefly. Even though the model's answer wasn't as detailed, it still nailed the key point about potential structural risk. Along with the earlier tasks (defect type, location, and shape), this example shows the model can handle tougher, more interpretive VQA questions. Its BLEU scores and alignment with expert insights suggest it's a strong tool for supporting automated structural evaluations in NDE work.

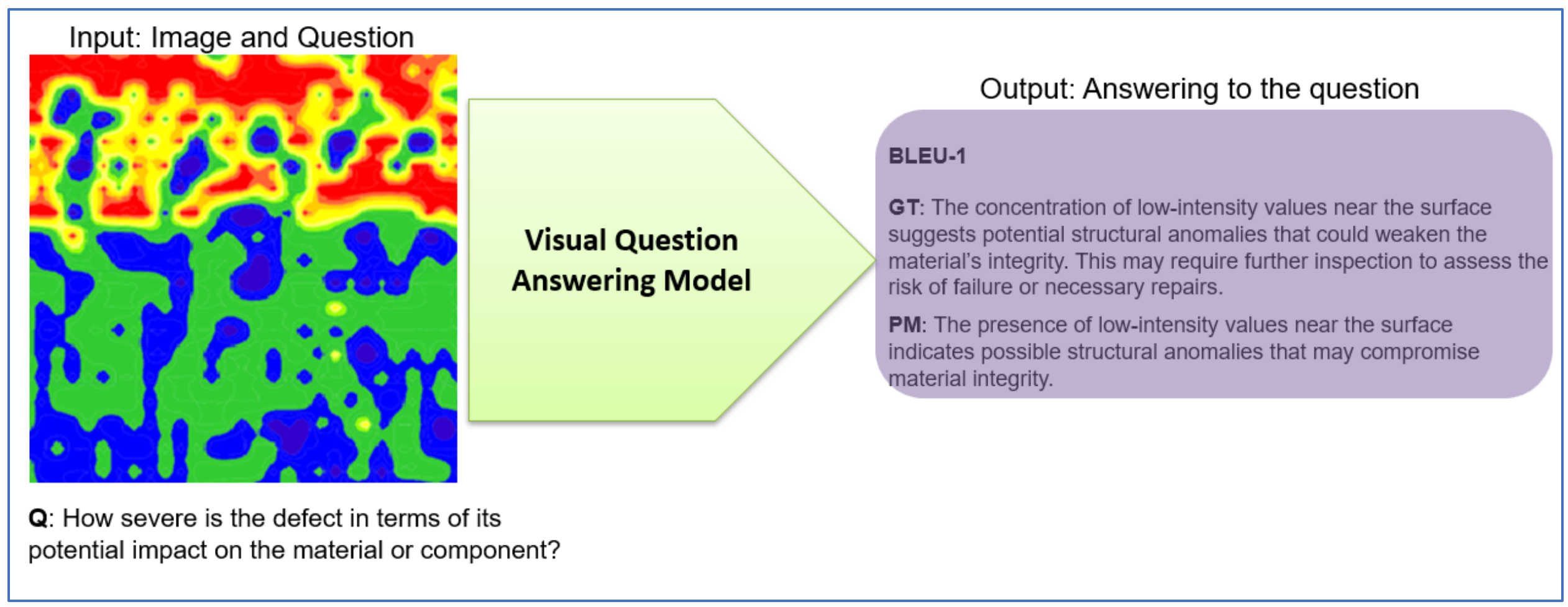

*Figure 17. The input for VQA model is an Impact Echo image paired with a domain-specific question: "How severe is the defect in terms of its potential impact on the material or component?"*

Also, this new round of VQA results on different IE images show that the model does a solid job answering a variety of defect-related questions based on Impact Echo images. In all four scenarios—figuring out the defect's type, location, shape, and severity (Figure 18)—the model's answers closely matched what the experts said. The BLEU-1 scores stayed high, which means the model consistently used the right terms and captured the meaning accurately. In the first two cases, it correctly identified the defect as delamination or voids and pinpointed its location in the upper-left area of the image, based on low-intensity values. For the third question about shape, it reasoned that delaminations usually show up as irregular or layered—proving it can handle trickier, less obvious visual traits. And in the last case, when asked about severity, the model connected those clustered low-response areas to potential structural weakness. All in all, these results highlight how reliable and flexible the model is when it comes to answering a wide range of domain-specific questions in NDE image analysis.

Input: Image and Question

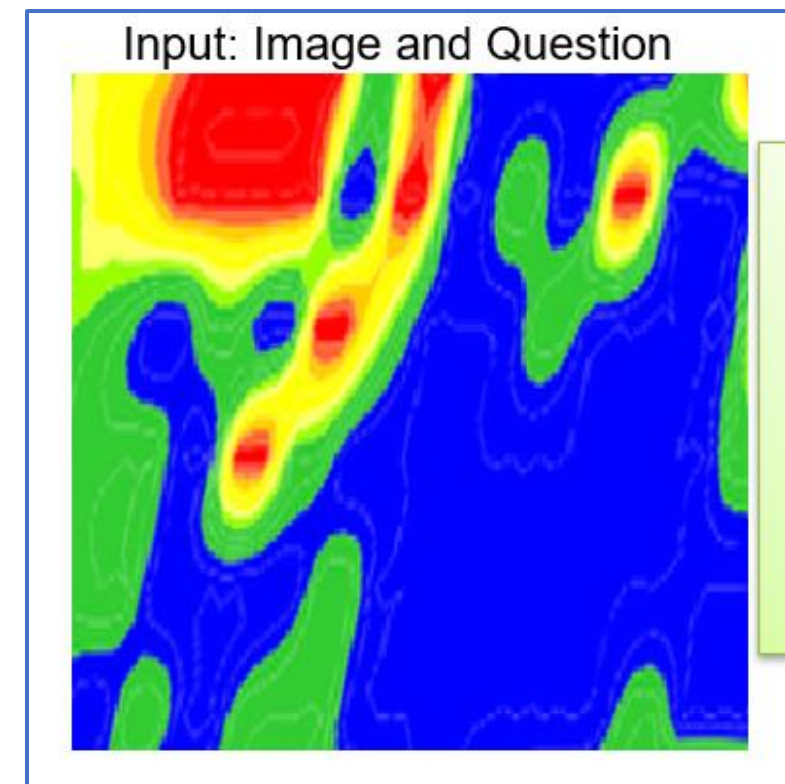

**Q**: What type of defect is observed in the image?

Visual Question Answering Model

Output: Answering to the question

**BLEU-1**

**GT**: The image indicates surface-level defects, such as voids or delamination, as identified in the low-response areas (red to yellow, values 1000–5000) in the upper-left section.

**PM**: The image reveals surface defects like delamination, seen in low-response in the upper-left area.

(a)

Input: Image and Question

**Q**: Where is the defect located within the material or component?

Visual Question Answering Model

Output: Answering to the question

**BLEU-1**

**GT**: The defects are concentrated in the upper-left section of the image, where low-response values (1000–5000) are present, suggesting areas of reduced material integrity.

**PM**: The defects are mainly located in the upper-left part of the image.

(b)

Input: Image and Question

**Q**: What is the shape of the defect—linear, circular, irregular, or another form?

Visual Question Answering Model

Output: Answering to the question

**BLEU-1**

**GT**: Seems a circular void and delaminations often appear as irregular or layered separations within the material. Further analysis would be needed to determine the exact shape.

**PM**: A circular void is possible, while delamination typically appears as irregular or layered separations in the material.

(c)

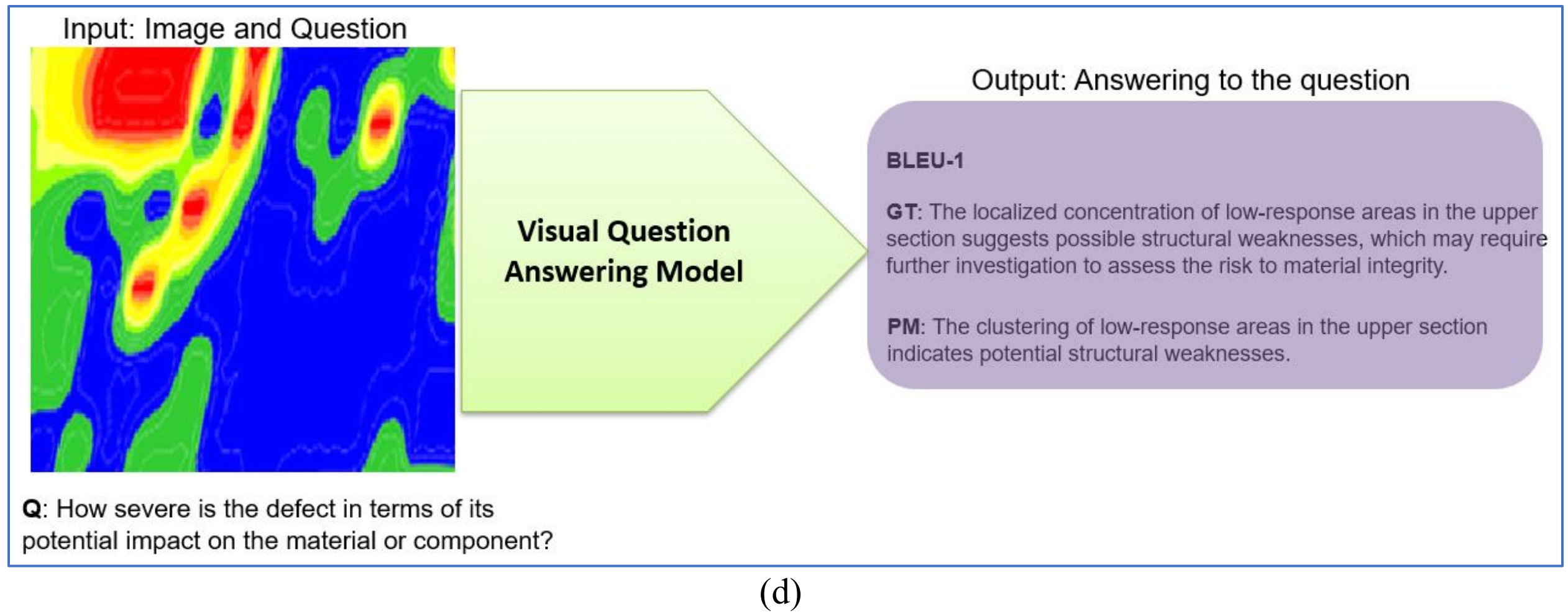

(d)

*Figure 18. The input for VQA model is an Impact Echo image paired with a domain-specific questions*

*4.5.2. GPR images*

The VQA model was also put to the test using Ground Penetrating Radar (GPR) data, where the focus was on interpreting attenuation maps by answering location-specific questions (Figure 19). In the first example, the model was asked to find the red and dark red regions—usually signs of serious deterioration—and it described it by correctly pointing to the lower left and center parts of the image, just like the ground truth. In the second example, the model was asked about the blue areas, and it correctly described them as showing up in the upper and central zones, linking them to minimal deterioration—again matching the expert answer. For the third question, it was asked about the yellow regions, and it accurately explained that they appear around red areas, indicating moderate deterioration. In all three cases, the model showed it could understand both the visual layout and the meaning of the colors in the context of deterioration. The BLEU-1 scores stayed high, which means the model consistently used the right terms and captured the meaning accurately (Table 5). Its answers were short but spot-on. These results show that the model doesn't just work well with Impact Echo data—it can handle GPR images too and give reliable answers across different types of questions.

*Table 5. Evaluation of the trained model using BLEU Matrix*

| Matrix | BLEU-1 | BLEU-2 |
|---|---|---|
| Model | 0.75 | 0.71 |

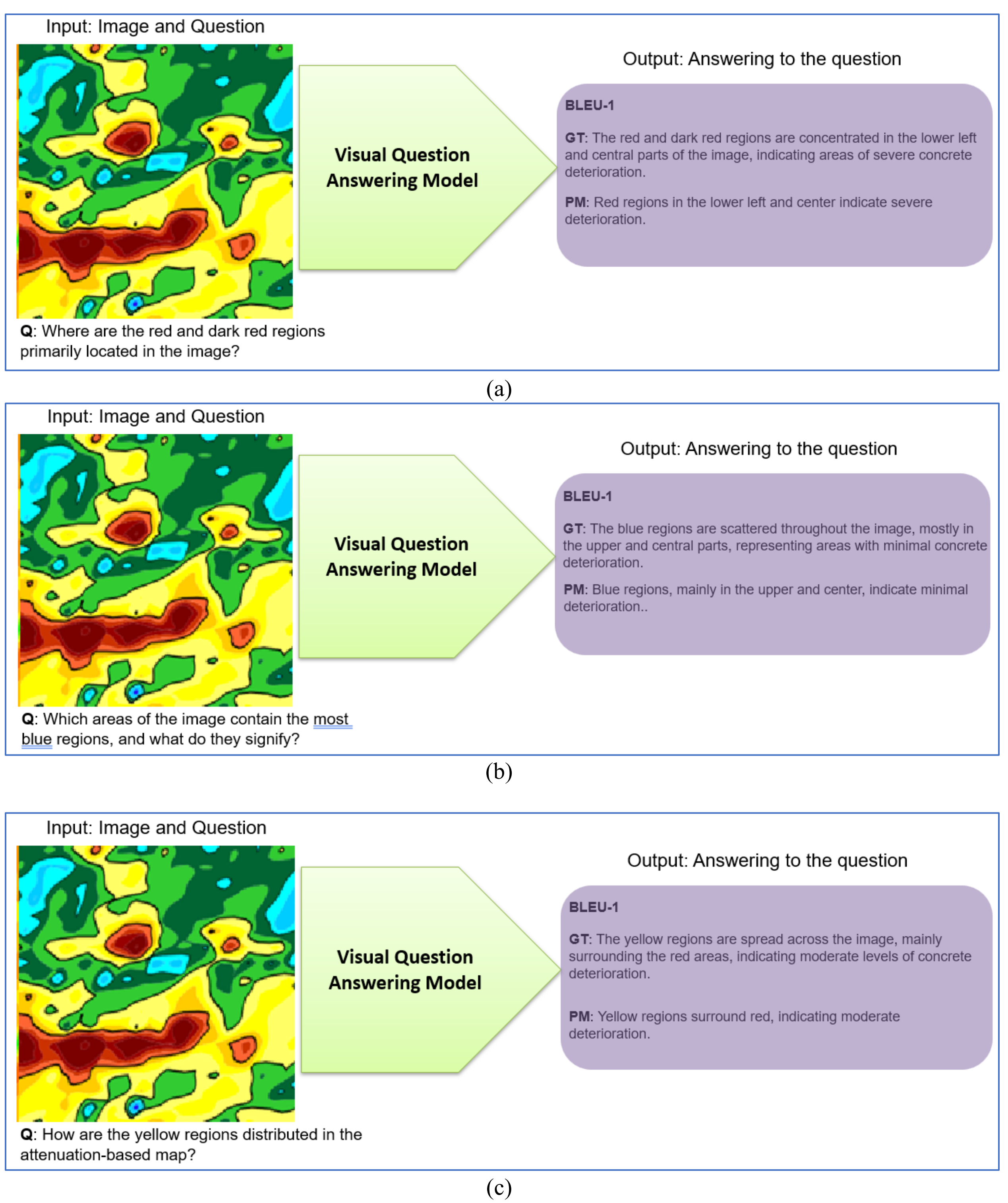

*Figure 19 The input for VQA model is an GPR image paired with a domain-specific questions.*

This new set of VQA results using a different Ground Penetrating Radar (GPR) dataset shows that the model still performs really well when it comes to reading attenuation maps. In the first example, it accurately pointed out that the red and dark red areas—usually signs of serious concrete damage—are mostly in the center of the image, which matched what the experts had noted. For the second question about the blue areas, the model correctly observed that there were few or none, which it interpreted as a sign of widespread deterioration—right on target. In the third case, when asked about the yellow regions, the model said they were mostly around the orange and red zones, which lined up nicely with the expert explanation that these areas show moderate deterioration. In all three cases, the model's answers were short but captured the key points, showing it understands both the visuals and the meaning behind them, even when working with different GPR data (Figure 20).

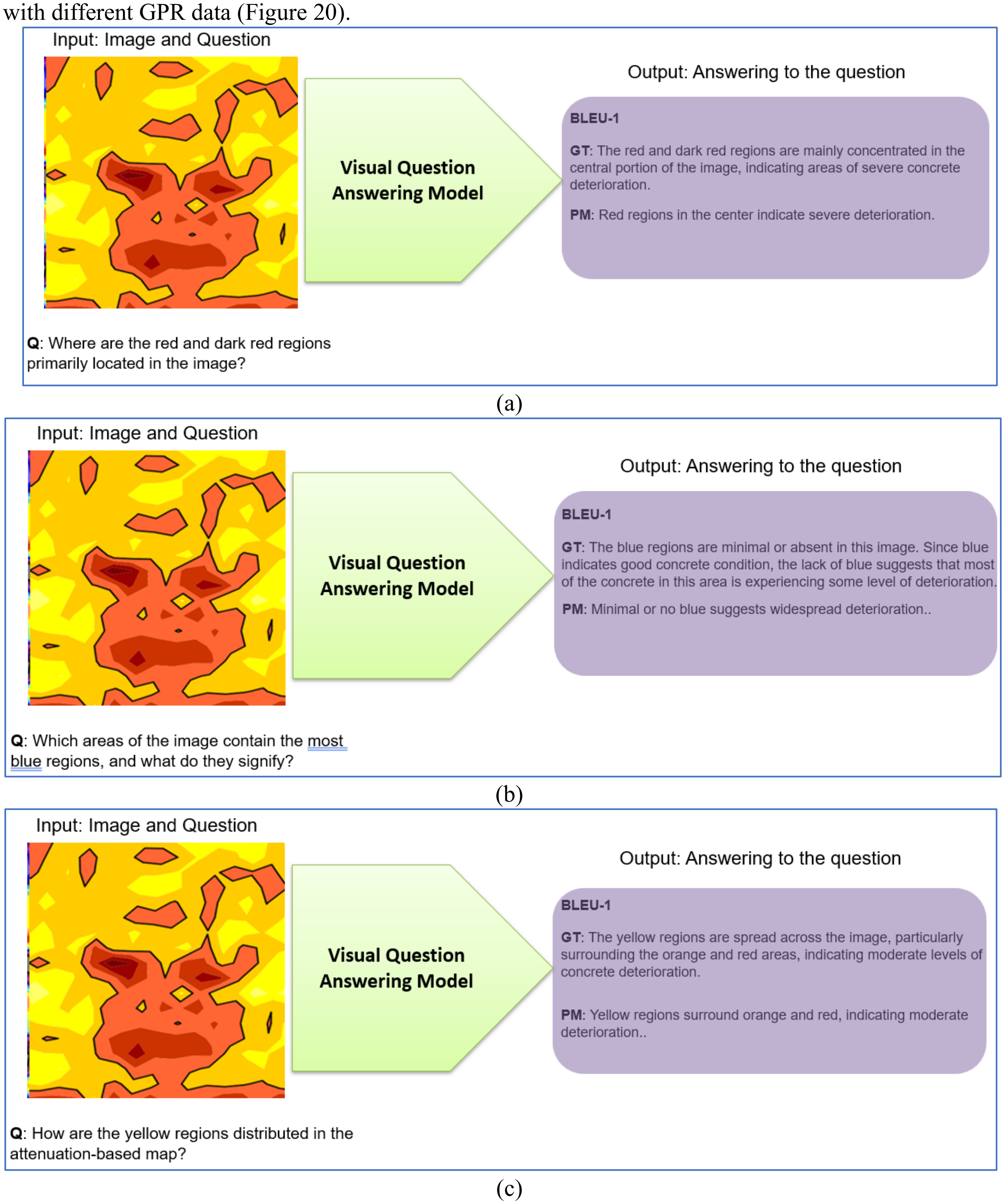

*Figure 20. The input for VQA model is an GPR image paired with a domain-specific questions.*

## 5. **Conclusion & Future Work**

The ChatNDE-Figure-to-Caption framework aims to automate Non-Destructive Evaluation (NDE) image analysis using AI, thereby improving efficiency and diagnostic accuracy. The proposed method integrates a ResNet-50 + GPT-2 architecture for caption generation, a Vision-Language Pretraining (VLP) model for image-caption understanding, and a Visual Question Answering (VQA) model for interpreting question–answer pairs. The BEAST dataset was used for training and validation across both captioning and VQA tasks. Initial results from the VLP model, which was trained on image–caption pairs from Impact Echo data, showed relatively low accuracy on test samples. The model particularly struggled with generating long, descriptive captions, often falling short in producing complete or detailed outputs. In contrast, the VQA model—fine-tuned on curated question–answer data—demonstrated much stronger performance. Using the BLEU metric for evaluation, the VQA model consistently produced accurate and concise responses to technical questions, especially for tasks involving defect type, location, shape, severity, and material conditions from both Impact Echo and GPR imagery. The improved accuracy is largely attributed to the shorter, focused nature of VQA responses, making it a more effective tool for real-world NDE applications.

To make the ChatNDE-Figure-to-Caption model even better and more useful, there are a few promising directions for future work. First off, boosting the captioning skills of the VLP model is a big priority. This could mean fine-tuning it on a larger, more diverse dataset that includes detailed annotations and longer, more descriptive captions—basically teaching the model to better understand technical language and complex structural details. Bringing in more advanced language models like GPT-4 or domain-specific transformers could also help improve the fluency and richness of the generated text. Next, the VQA dataset could be expanded with more variety in question types—like multi-step reasoning, time-based context, or questions that ask the model to show confidence in its answers. This would make the model more capable of handling tough, real-world decision-making scenarios. Finally, combining multiple types of NDE data—like Impact Echo, GPR, Electrical Resistivity, and Ultrasonic Surface Waves—could give a more complete picture of what's going on below the surface. This kind of data fusion could be powered by graph-based models or spatiotemporal frameworks that are built to handle multiple data types at once.

**Acknowledgements**

The authors gratefully acknowledge the BEAST facility management team and Rutgers University for providing access to the NDE data. The opinions, findings, conclusions, and recommendations presented in this publication are solely those of the author(s) and do not necessarily represent or reflect the official views of the Federal Highway Administration (FHWA).

**Competing Interests**

The authors declare no competing interests.

**APPENDEX**

This script trains an AI model that can look at an image and generate a caption describing what’s in it. It starts by loading a dataset of images and their matching captions from a JSON file. The images are preprocessed (resized, cropped, normalized) so they’re ready for the model. For the vision part, it uses a ResNet-50 model—a popular deep learning model for recognizing image features. But instead of using it for classification, the script removes the final layer so it just extracts visual features from the image. Then for the language part, it uses GPT-2, which is a well-known model for generating text. But GPT-2 normally expects words, not images—so the code adds a linear layer that maps the image features into a format GPT-2 understands. During training, it combines the image features with a short prompt (like "An image showing...") and teaches GPT-2 to finish the sentence with a caption that matches the actual one from the dataset. It uses the BLEU loss built into GPT-2 to measure how close its output is to the real caption. The model gets trained for a bunch of epochs (up to 1000 in this case), and after training, it saves the model and tries it out by generating captions for new images. It even shows the image with the generated caption using matplotlib so you can see how well it did.

```
1  import torch
2  import torch.nn as nn
3  import torchvision.models as models
4  from torchvision import transforms
5  from torch.utils.data import Dataset, DataLoader
6  from PIL import Image
7  import json
8  from transformers import GPT2Tokenizer, GPT2LMHeadModel, AdamW
9  import matplotlib.pyplot as plt
10
11 # Define your image caption dataset class
12 class ImageCaptionDataset(Dataset):
13     def __init__(self, json_file, transform=None):
14         with open(json_file, 'r') as file:
15             self.dataset = json.load(file)
16         self.transform = transform
17
18     def __len__(self):
19         return len(self.dataset)
20
21     def __getitem__(self, idx):
22         item = self.dataset[idx]
23         image = Image.open(item['image_path']).convert("RGB")  # Ensure RGB format
24         if self.transform:
25             image = self.transform(image)
26         caption = item['caption']
27         return image, caption
28
29 # Define transformations for your images
30 transform = transforms.Compose([
31     transforms.Resize(256),          # Resize shorter side to 256 pixels
32     transforms.CenterCrop(224),      # Center crop to 224x224 pixels
33     transforms.ToTensor(),           # Convert image to PyTorch tensor
34     transforms.Normalize(mean=[0.485, 0.456, 0.406], std=[0.229, 0.224, 0.225]),  # Normalize with ImageNet mean and std
35 ])
36
```

```
# Load the dataset
json_file_path = "C:/Users/m.shafiei.dizaji.ctr/Desktop/dataset/captions.json"  # Path to your JSON file
dataset = ImageCaptionDataset(json_file=json_file_path, transform=transform)
dataloader = DataLoader(dataset, batch_size=2, shuffle=True)  # Adjust batch size as needed

# Load pre-trained ResNet model for feature extraction
resnet = models.resnet50(pretrained=True)
resnet.eval()

# Modify the ResNet model to remove the final classification layer
class ResNetFeatureExtractor(nn.Module):
    def __init__(self):
        super(ResNetFeatureExtractor, self).__init__()
        self.resnet = nn.Sequential(*list(resnet.children())[:-1])  # Remove the last layer

    def forward(self, x):
        with torch.no_grad():
            features = self.resnet(x)
        return features.squeeze()

feature_extractor = ResNetFeatureExtractor()

# Load pre-trained GPT-2 model and tokenizer for caption generation
tokenizer = GPT2Tokenizer.from_pretrained("gpt2")
tokenizer.pad_token = tokenizer.eos_token  # Set eos_token as pad_token
gpt2 = GPT2LMHeadModel.from_pretrained("gpt2")
gpt2.eval()

# Define a linear layer to map the visual features to the text space
class ImageCaptioningModel(nn.Module):
    def __init__(self):
        super(ImageCaptioningModel, self).__init__()
        self.feature_extractor = feature_extractor
        self.feature_to_embedding = nn.Linear(2048, gpt2.config.n_embd)  # Map features to GPT-2 embedding size
        self.gpt2 = gpt2

    def forward(self, images, input_ids, labels=None):
        features = self.feature_extractor(images)
        embeddings = self.feature_to_embedding(features).unsqueeze(1).expand(-1, input_ids.size(1), -1)
        gpt2_outputs = self.gpt2(inputs_embeds=embeddings, labels=labels)
        return gpt2_outputs

# Instantiate the image captioning model
model = ImageCaptioningModel()
model.train()  # Set model to training mode

# Define the optimizer
optimizer = AdamW(model.parameters(), lr=5e-5)

# Training parameters
num_epochs = 1000
loss_values = []  # List to store loss values

# Training loop
for epoch in range(num_epochs):
    epoch_loss = 0  # Initialize loss for the epoch
    for images, captions in dataloader:
        # Ensure the batch is not empty
        if images.size(0) == 0:
            print("Skipping empty batch.")
            continue

        # Prepare input text and labels
        input_text = "An image showing a detailed scene with multiple elements, including people, objects, and various activities."
        input_ids = tokenizer(input_text, return_tensors="pt", padding=True, truncation=True, max_length=20).input_ids
        input_ids = input_ids.repeat(images.size(0), 1)

        # Prepare the target text
        target_texts = [f"{input_text} {caption}" for caption in captions]
        labels = tokenizer(target_texts, return_tensors="pt", padding=True, truncation=True, max_length=20).input_ids

        # Ensure labels are the same size as input_ids by padding
        labels = labels[:, :input_ids.size(1)]
        if labels.size(1) < input_ids.size(1):
            labels = torch.nn.functional.pad(labels, (0, input_ids.size(1) - labels.size(1)), value=tokenizer.pad_token_id)
```

```
        # Zero the gradients
        optimizer.zero_grad()

        # Forward pass
        outputs = model(images, input_ids, labels=labels)
        loss = outputs.loss

        # Backward pass and optimization
        loss.backward()
        optimizer.step()

        epoch_loss += loss.item()  # Accumulate the loss for the epoch

    # Store the average loss for the epoch
    loss_values.append(epoch_loss / len(dataloader))

    print(f"Epoch [{epoch + 1}/{num_epochs}] completed with loss: {epoch_loss / len(dataloader)}")

# Save the trained model
model_save_path = "fine_tuned_image_captioning_model.pt"
torch.save(model.state_dict(), model_save_path)
print(f"Model saved to {model_save_path}")

# Plot loss versus epochs
plt.plot(range(1, num_epochs + 1), loss_values, marker='o')
plt.xlabel('Epoch')
plt.ylabel('Loss')
plt.title('Training Loss vs Epochs')
plt.grid(True)
plt.show()

# Function to generate caption and display image
def display_image_with_caption(image_tensor, caption):
    # Convert the tensor to a PIL image
    image = transforms.ToPILImage()(image_tensor.cpu().squeeze(0))

    # Display the image with its caption
    plt.imshow(image)
    plt.title(caption)
    plt.axis('off')
    plt.show()

# Example of generating captions for a batch of images after training
model.eval()  # Set model to evaluation mode
for images, _ in dataloader:
    # Print shapes to debug
    print(f"Images shape: {images.shape}")

    # Ensure the batch is not empty
    if images.size(0) == 0:
        print("Skipping empty batch.")
        continue

    # Tokenize the initial text (e.g., "An image showing a detailed scene with multiple elements") and prepare input IDs
    input_text = "An image showing a detailed scene with multiple elements, including people, objects, and various activities."
    input_ids = tokenizer(input_text, return_tensors="pt", padding=True, truncation=True, max_length=20).input_ids

    # Ensure input_ids batch size matches images batch size
    input_ids = input_ids.repeat(images.size(0), 1)

    # Generate the caption
    outputs = model(images, input_ids)
    generated_ids = outputs.logits.argmax(-1)
    generated_text = tokenizer.decode(generated_ids[0], skip_special_tokens=True)

    print(f"Generated Caption: {generated_text}")

    # Display the image with its generated caption
    display_image_with_caption(images[0], generated_text)
    break  # Just one batch for demonstration
```

## Another example for VLP model

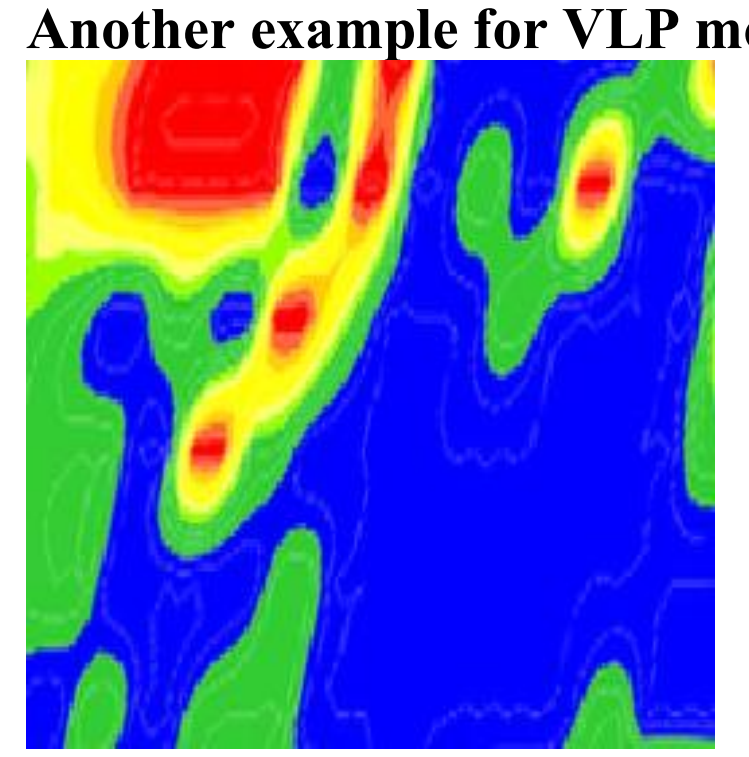

**GT**: *"Impact Echo (IE) image illustrating subsurface response variations, with color intensities corresponding to response values. Low-response areas (in red to yellow, values around 1000–5000) are concentrated in the upper-left section, suggesting potential surface-level defects, such as voids or delaminations, indicating reduced material integrity in these regions. Intermediate-response zones (in green, values around 5000–7000) are observed along the edges and near the lower-left, representing areas with partial material consistency. High-response regions (in blue, values around 9000–10000) dominate the lower portion of the image, indicating intact material with strong, uniform properties at greater depths. The localized concentration of low-response areas in the upper section suggests possible structural weaknesses that may need further investigation."*

**PM**: *Impact Echo image shows subsurface variations. Low-response suggests surface defects, medium indicates partial consistency, and high signals strong material. Clusters of low response may indicate structural issues*